\PassOptionsToPackage{hyphens}{url}  
\documentclass[lettersize,journal]{IEEEtran}
\usepackage{amsmath,amsfonts}
\usepackage{algorithmic}
\usepackage{algorithm}
\usepackage{array}
\usepackage[caption=false,font=normalsize,labelfont=sf,textfont=sf]{subfig}
\usepackage{textcomp}
\usepackage{stfloats}
\usepackage{url}
\usepackage{verbatim}
\usepackage{graphicx}
\usepackage{cite}
\usepackage{pifont}
\usepackage[table,xcdraw]{xcolor}

\usepackage{makecell}

\usepackage{pgfplots}
\usepackage{hyperref}
\usepackage{ragged2e}

\hypersetup{
colorlinks=true,
citecolor=green,
citebordercolor=green
}

\begin{document}
\title{Bridging Perspectives:\\ A Survey on Cross-view Collaborative Intelligence with Egocentric-Exocentric Vision}

\author{Yuping He, Yifei Huang, Guo Chen, Lidong Lu, Baoqi Pei, Jilan Xu,\\Tong Lu, Yoichi Sato
\thanks{

}
\thanks{\IEEEcompsocthanksitem Y. He, Y. Huang, and G. Chen have equal contributions. Y. He, G. Chen, L. Lu and T. Lu are with the State Key Laboratory for Novel Software Technology, Nanjing University, Nanjing 210023, China. Y. Huang and Y. Sato are with the University of Tokyo, Tokyo, Japan. B. Pei is with Zhejiang University, Zhejiang 310027, China. J. Xu is with Fudan University, Shanghai 200433, China. }

}



\maketitle

\begin{abstract}
Perceiving the world from both egocentric (first-person) and exocentric (third-person) perspectives is fundamental to human cognition, enabling rich and complementary understanding of dynamic environments. In recent years, allowing the machines to leverage the synergistic potential of these dual perspectives has emerged as a compelling research direction in video understanding. In this survey, we provide a comprehensive review of video understanding from both exocentric and egocentric viewpoints. We begin by highlighting the practical applications of integrating egocentric and exocentric techniques, envisioning their potential collaboration across domains. We then identify key research tasks to realize these applications. Next, we systematically organize and review recent advancements into three main research directions: (1) leveraging egocentric data to enhance exocentric understanding, (2) utilizing exocentric data to improve egocentric analysis, and (3) joint learning frameworks that unify both perspectives. For each direction, we analyze a diverse set of tasks and relevant works. Additionally, we discuss benchmark datasets that support research in both perspectives, evaluating their scope, diversity, and applicability. Finally, we discuss limitations in current works and propose promising future research directions. By synthesizing insights from both perspectives, our goal is to inspire advancements in video understanding and artificial intelligence, bringing machines closer to perceiving the world in a human-like manner. A GitHub repo of related works can be found at \url{https://github.com/ayiyayi/Awesome-Egocentric-and-Exocentric-Vision}.
\end{abstract}


\section{Introduction}


\IEEEPARstart{P}{erceiving} the world from both egocentric (first-person) and exocentric (third-person) perspectives is a fundamental ability in human intelligence. The mirror neuron theory \cite{mirror} posits that the same neural mechanisms are activated when an individual performs an action and when they observe another performing the same action. This biological insight underscores the intrinsic connection between first- and third-person viewpoints, inspiring efforts to emulate this capability. By enabling machines to integrate and leverage information across these perspectives, we can advance video understanding and move closer to human-like perception.

The exocentric (third-person) and egocentric (first-person) perspectives offer complementary views of human activity, akin to two sides of the same coin. The egocentric view provides an actor-centered perspective \cite{actorobserver}, capturing rich human-object interactions and reflecting the wearer’s intentions and goals~\cite{epic, ego4d, goalstep}. Unlike the exocentric view, egocentric videos are inherently more dynamic, featuring continuous motion and shifting backgrounds, which pose challenges such as partial visibility of the wearer~\cite{egohumans, Fan_2017_CVPR}. Still, the release of large-scale egocentric datasets~\cite{epic, ego4d, Egoexo4d, hdepic} has spurred substantial progress in egocentric video understanding~\cite{egovlp, egovlpv2, Helping_Hands, chen2022internvideo-ego4d,pei2024egovideo,pei2025modeling}.

In contrast, the exocentric view offers an observer-like perspective \cite{actorobserver}, providing a broader context of the scene and the subject’s actions. Different from egocentric videos, these videos are usually recorded from a stable, fixed position, covering a wide field of view and capturing detailed scene context. These videos can be easily captured using devices such as smartphones and surveillance cameras, and their widespread availability on the Internet has led to the creation of diverse large-scale datasets, for example, \cite{HowTo100M, ava, Soomro2012UCF101AD, kinetic,wang2023internvid,chen2024cg-bench}. These datasets have driven significant advancements in third-person video understanding \cite{Two_streamCNN,kinetic,vivit,videotransformer,chen2022dcan,chen2024video-mamba-suite,wang2023mat}.

\begin{figure}[t]
\centering
\includegraphics[width=1.0\columnwidth]{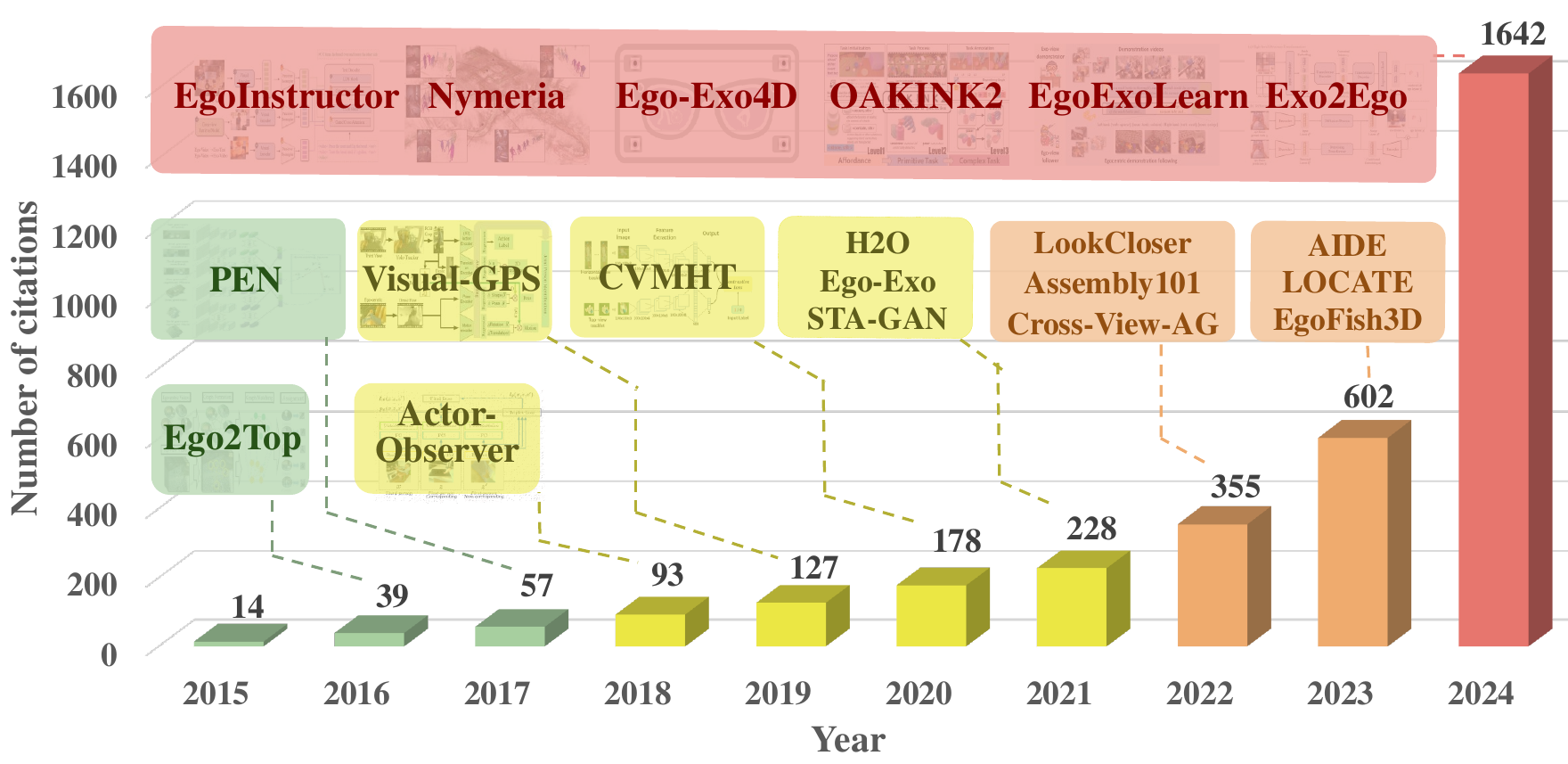}
\caption{Number of citations to egocentric-exocentric related papers from 2015 to 2024. Citation data was collected from Google Scholar. The statistics are computed based on papers and datasets discussed in Sections \ref{section: task} and \ref{section:dataset}, all of which utilize both egocentric and exocentric perspectives. }
\label{fig:citation}
\vspace{-6mm}
\end{figure}

\IEEEpubidadjcol

While egocentric and exocentric perspectives have distinct characteristics, they are inherently complementary~\cite{Egoexo4d}.
The ego-view provides details from the actor’s perspective, while the exo-view offers a broader contextual understanding of the scene. Researchers can unlock new opportunities to advance video understanding by integrating these perspectives. This synergy has led to a growing body of work exploring cross-view learning, as demonstrated in Fig. \ref{fig:citation}.

Despite these advancements, there remains a lack of surveys that summarize progress in integrating both perspectives. In video understanding, most surveys \cite{vsr_survey, action_survey, detection_survey, anomaly_survey} focus on specific tasks and primarily concentrate on exocentric videos. In egocentric vision, Plizzari \textit{et al.}~\cite{plizzari2024outlook} review advancements across multiple tasks. However, to the best of our knowledge, no survey has yet addressed the integration of both perspectives. 

Thus, our work fills this gap by systematically organizing and reviewing existing research into three primary directions: (1) leveraging egocentric data to enhance exocentric understanding, (2) utilizing exocentric data to improve egocentric analysis, and (3) joint learning frameworks for cross-view video understanding.

\begin{figure*}[h]
\centering
\includegraphics[width=1.0\textwidth]{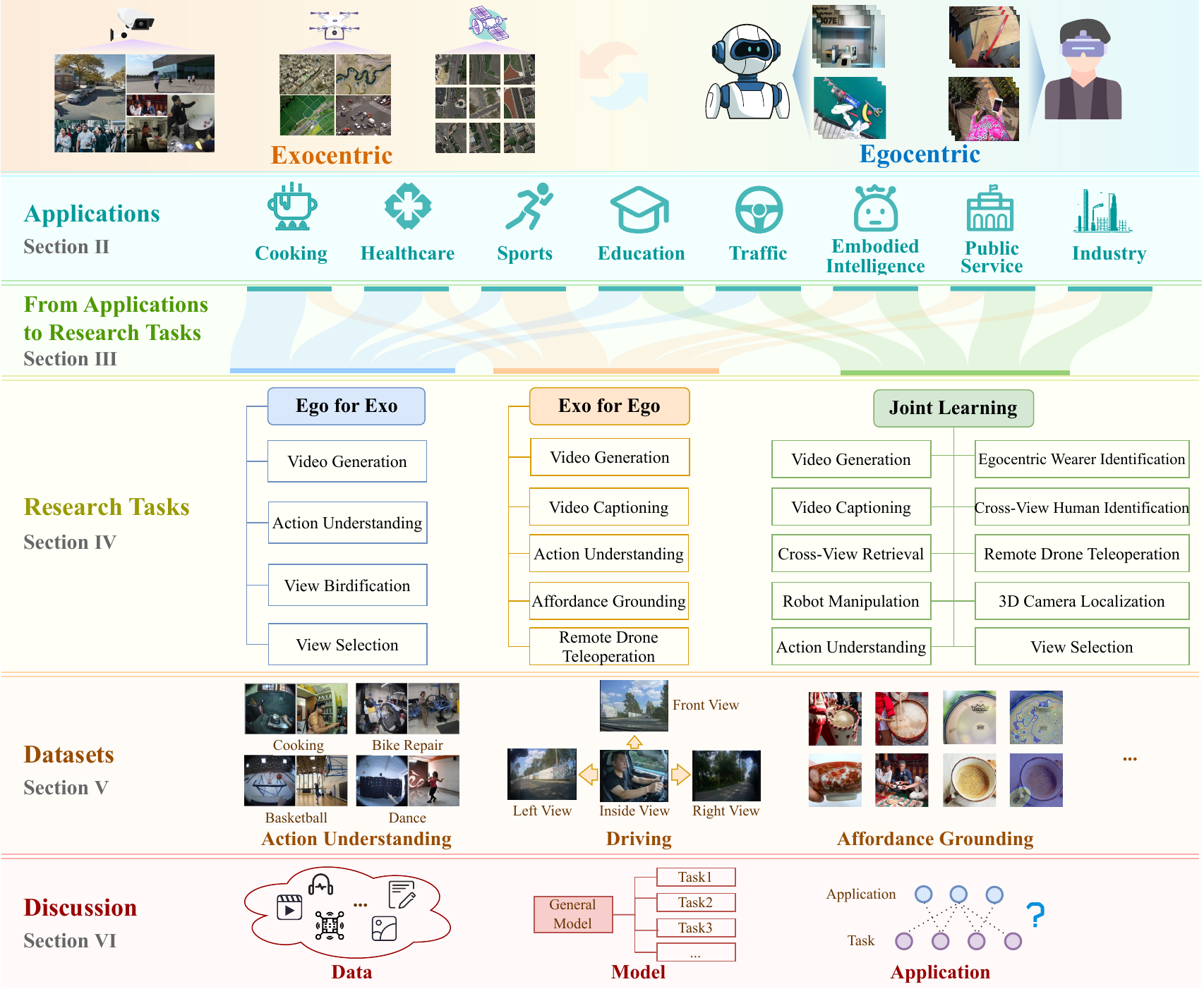}
\caption{The overall structure of the survey. We first highlight the application value of egocentric and exocentric collaboration (Section \ref{section: application}). We then identify critical research tasks for each application (Section \ref{section: mapping}). Next, we provide a comprehensive overview of the current research advancements (Section \ref{section: task}). This section is divided into: ego for exo, exo for exo, and joint learning, each covering various research tasks. Additionally, we examine datasets that encompass both perspectives (Section \ref{section:dataset}). Finally, we discuss limitations and future directions (Section \ref{section:outlook}).}
\label{fig:structure}
\vspace{-4mm}
\end{figure*}

The overall structure of this survey is illustrated in Fig.~\ref{fig:structure}. 
Inspired by \cite{plizzari2024outlook}, we also adopt a ``future-to-present" approach. Specifically, we start by highlighting the transformative potential of integrating egocentric and exocentric perspectives~\cite{Egoexo4d}, demonstrating how cross-view collaboration can benefit various domains (Section \ref{section: application}). 
We then identify key research tasks to realize these applications (Section \ref{section: mapping}). 
In addition to the systematic review of existing research works (Section \ref{section: task}), we also analyze benchmark datasets that support both perspectives (Section \ref{section:dataset}), evaluate their diversity and applicability. Finally, we discuss the limitations of current approaches and propose promising research directions (Section \ref{section:outlook}).

\section{Applications} \label{section: application}
\begin{figure*}[t]
\centering
\includegraphics[width=2.0\columnwidth]{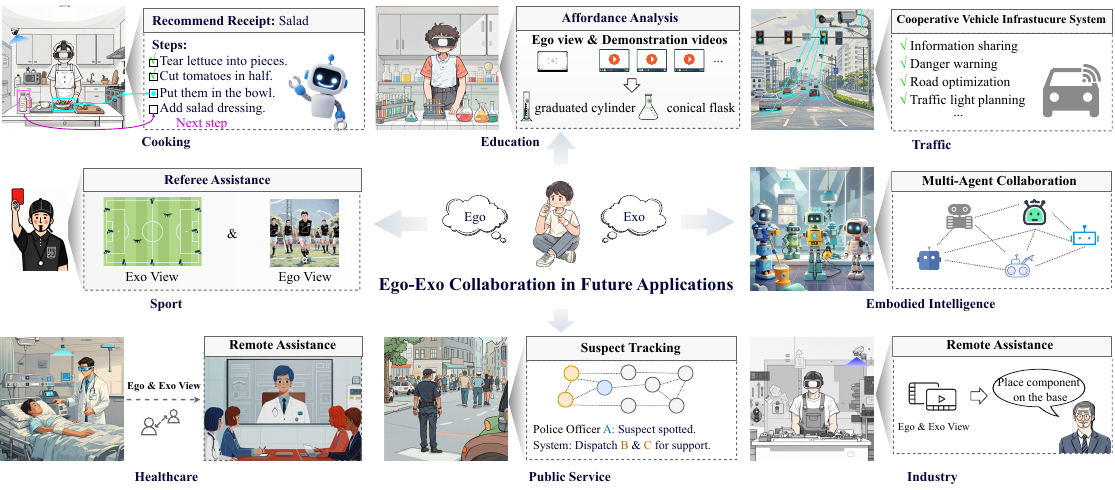}
\caption{Examples of the potential collaboration of egocentric and exocentric vision in diverse applications. We illustrate how integrating egocentric and exocentric video understanding techniques can enhance these applications.}
\label{fig:envision}
\vspace{-5mm}
\end{figure*}

In this section, we highlight the practical value of egocentric and exocentric video understanding techniques. We select eight representative application scenarios that have a significant demand for ego-exo collaboration. For each scenario, we provide examples of how egocentric or exocentric techniques are applied in real-world systems. Notably, most current applications are limited to a single perspective. Therefore, we explore how ego-exo collaboration could drive future innovations, as demonstrated in Fig.~\ref{fig:envision}.

\vspace{-1em}
\subsection{Cooking}
Vision-based kitchen assistants have recently emerged, with systems like the Samsung Family Hub refrigerator \cite{nick-lavars_2016} and the June Oven \cite{june_oven_2018} using exocentric cameras for food recognition and task-specific automation. However, these systems are limited in scope and lack holistic cooking support.

In future kitchens, we imagine exocentric cameras will work with head-mounted AR glasses to assist cooking. The head-mounted camera will identify ingredients and their freshness, recommend items, and display them in the AR glasses. 
During cooking, the AR glasses will recognize current steps and display the next step, while overhead cameras will monitor the workspace to prevent accidents. Thus, techniques like ego-exo action recognition and cross-view associations of key steps, can greatly help these kitchen applications.

\vspace{-1em}
\subsection{Sports}

Exocentric vision systems currently dominate sports analysis, with applications such as sports tracking systems~\cite{mcdonald_2013} and referee assistance system~\cite{winter_2024}. For broadcasting, Fox Sports' ``Be The Player"~\cite{dachman_2017} generates egocentric replays from exocentric views. However, as wearing cameras can hinder players’ movements, the use of egocentric perspectives and multi-view collaboration remains limited.

In the future, advancements in wearable technology will enable lightweight egocentric devices tailored for athletes. These devices can capture fine-grained details of athletes’ movements. For referees, integrating multi-view video footage can enhance decision-making. Realizing this multi-perspective approach requires techniques like cross-view person identification and tracking for seamless cross-view data alignment.

\vspace{-0.5em}
\subsection{Healthcare}

Currently, exocentric cameras are extensively deployed in hospitals, providing real-time observations of patients' health conditions. Besides, egocentric cameras worn by on-site doctors enable remote assistance \cite{asia_2022_smart} and emergency services~\cite{asia_2022_heres}. However, most current applications rely on a single view and lack multi-view collaboration.

For the future, integrating both views can enhance future medical practices. Remote experts can utilize both the surgeon's egocentric view and the exocentric recording cameras to give effective guidance. 
Similarly, remote therapists can benefit from multi-view data to give personalized care plans. These applications necessitate techniques such as ego-exo action assessment and pose estimation.

\vspace{-0.5em}
\subsection{Education}
Nowadays, cameras are widely installed on classroom ceilings to track student movements and enhance safety~\cite{classroom_cam_2}. Additionally, class recording systems capture lectures, supporting both review sessions and online learning~\cite{classroom_cam}. However, these systems currently operate as passive recording tools, lacking the ability to actively contribute to teaching activities.

Future intelligent classrooms will leverage egocentric and exocentric video collaboration for enhanced learning experiences. During laboratory sessions, egocentric cameras can complement exocentric demonstrations to teach unfamiliar instruments. Besides, transforming exocentric demonstrations into egocentric perspectives enhances intuitive learning. Thus, techniques like ego-exo affordance analysis and cross-view transformation will be key to personalized educational service. 

\vspace{-0.5em}
\subsection{Traffic}
Currently, onboard cameras are widely employed in driving assistance \cite{news_2018,freightwaves_2019}, and autonomous driving \cite{liu_2023} systems. Traffic management systems utilize surveillance cameras to monitor intersections to control traffic signals adaptively. However, data from onboard cameras and surveillance systems often lack coordination, limiting their combined potential.

Future traffic systems will enable information sharing between vehicles and road infrastructure. Onboard cameras will combine with the surveillance network to monitor the driver's state and enhance scene awareness. This requires techniques such as ego-exo action recognition and cross-view semantic segmentation. Additionally, vehicle footage will be uploaded to the cloud and combined with surveillance footage to optimize traffic management, making cross-view object identification required to track vehicles across videos.  

\vspace{-0.5em}
\subsection{Embodied Intelligence}

Modern robots leverage both egocentric and exocentric vision for diverse applications, including space exploration, medical assistance, customer service, and security~\cite{robot_news}. They can also learn from human demonstrations by mapping exocentric instructional videos onto their own egocentric views for skill acquisition. 
Looking ahead, multi-agent robotic systems will increasingly depend on cross-perspective collaboration. Estimating egocentric camera positions within a global exocentric frame will enhance coordination, while combining views across robots will enable accurate 3D scene reconstruction for improved situational awareness. These advancements require progress in ego-exo localization, multi-view reconstruction, and collaborative perception.

\vspace{-0.5em}
\subsection{Public Service}
Egocentric and exocentric videos play an essential role in public services. Surveillance cameras aid in locating criminals and missing persons, while body-worn cameras capture on-site scenes for law enforcement~\cite{nij_2022_over}. In search and rescue, aerial drone footage complements ground-level views for timely response~\cite{singh_2023_over}. However, these systems typically operate in isolation, limiting their effectiveness.
Future urban systems will benefit from integrating egocentric footage with surveillance networks. For instance, in suspect tracking, the police system uses data from the egocentric cameras on the policemen and street surveillance to track suspects and dispatch forces accordingly. To achieve this, cross-view human identification and egocentric wearer identification are essential for associating individuals across multiple perspectives. 

\vspace{-0.5em}
\subsection{Industry}

In modern manufacturing, ceiling-mounted cameras are widely employed for safety monitoring \cite{fogsphere_2023}. On automated assembly lines, cameras on robotic arms help precisely locate and assemble parts~\cite{begg_2024}. During quality inspection, multi-view scans accurately identify product defects~\cite{jobit_2024}. However, current industrial vision systems operate largely in isolated viewpoints, limiting their ability to provide comprehensive process monitoring.

Future smart factories will integrate wearable and fixed cameras for real-time process optimization and worker support. Overhead cameras can capture the overall workflow, while egocentric cameras track individual worker actions to identify inefficiencies. When operators encounter issues, real-time video streaming from both perspectives can facilitate remote troubleshooting by experts. Enabling these applications will require advancements in cross-view action assessment and multi-view scene understanding.

\section{From Applications To Research Tasks} \label{section: mapping}
The previous section outlines how egocentric and exocentric perspectives can collaborate to enable a wide range of applications. However, realizing these envisioned applications requires addressing several fundamental research challenges. In this section, we identify key research tasks that demand egocentric-exocentric collaboration and review existing efforts that contribute to their development.

We categorize these tasks from three directions: (1) Exocentric for Egocentric, leveraging exocentric knowledge to enhance egocentric video understanding; (2) Egocentric for Exocentric, utilizing egocentric cues to improve exocentric tasks; and (3) Joint Learning, which integrates both perspectives for cross-view understanding. 

\textbf{Cooking.} Analyzing human actions is crucial for providing personalized cooking guidance. Recent research explores how exocentric knowledge can improve egocentric action recognition \cite{li2021ego, truong2023cross, dou2024unlocking, holographic, quattrocchi2023synchronization} and how joint learning of representations from both perspectives enhances overall action modeling \cite{2014ar}. Additionally, transforming exocentric cooking videos into egocentric perspectives has been shown to enhance the immersive experience \cite{Exo2Ego-V, Cheng20244DIFF3D}. Furthermore, exocentric data also proves beneficial for improving egocentric video captioning \cite{ohkawa2023exo2egodvc, xu2024retrieval}, facilitating the summarization of cooking procedures.

\textbf{Sports.} Analyzing dynamic actions in sports is critical for skill assessment and injury prevention. Several studies \cite{wang2022estimating, dhamanaskar2023enhancing} enhance egocentric pose estimation with exocentric data, while others focus on cross-view action recognition \cite{anegolook, AE2}. Moreover, transforming exocentric sports video into egocentric viewpoints can provide immersive training experiences, which has been investigated in basketball scenarios \cite{Exo2Ego-V, Cheng20244DIFF3D}.

\textbf{Healthcare.} Egocentric perspectives play a crucial role in medical training and remote assistance. Exocentric-to-egocentric transformations have been explored for procedural skill acquisition, including COVID testing and CPR \cite{Exo2Ego-V, Cheng20244DIFF3D}. In surgical environments, the doctor's first-person views can be utilized to select best view for recording system \cite{Saito}. Additionally, multi-view setups improve pose estimation of surgical instruments \cite{hein2023nextgeneration}, facilitating precise tool manipulation.

\textbf{Traffic.} Monitoring driver behavior is a key component of driver monitoring systems to enhance safety. As discussed in \cite{lbw, yang2023aide}, both in-vehicle and out-vehicle view are essential to recognize the driver's condition. 

\textbf{Embodied Intelligence}. For robotic manipulation, multi-view settings enables precise control \cite{Hsu2022VisionBasedMN, Jangir, shang, Acar, Sharma2019, dunion2024multiview, MV-MWM, MFSC}. Additionally, affordance grounding helps robots learn to use tools \cite{zhang2024self, yang2024learning}. Transforming exocentric demonstration videos into the robot’s view facilitates imitation learning \cite{garello2022towards, spisak2024diffusing}. Moreover, the transformed exocentric view can address the limited egocentric view of submersible vehicles \cite{water}. Robots can also act as valuable assistants in human-drone collaboration~\cite{Morando}, exocentric camera registration \cite{YOWO} and lifelog video captioning \cite{lifelog}.

\textbf{Industry.} Affordance grounding assists robots to use tools. In \cite{yang2024learning}, this task is extended to predict tool-based grasping regions. In technical training, converting exocentric demonstrations to first-person perspectives helps workers visualize procedural steps from their own views. This task has been explored in bike repair \cite{Exo2Ego-V, Cheng20244DIFF3D} and assembly \cite{luo2025put} scenarios.

\begin{figure}[t]
\centering
\includegraphics[width=1.0\columnwidth]{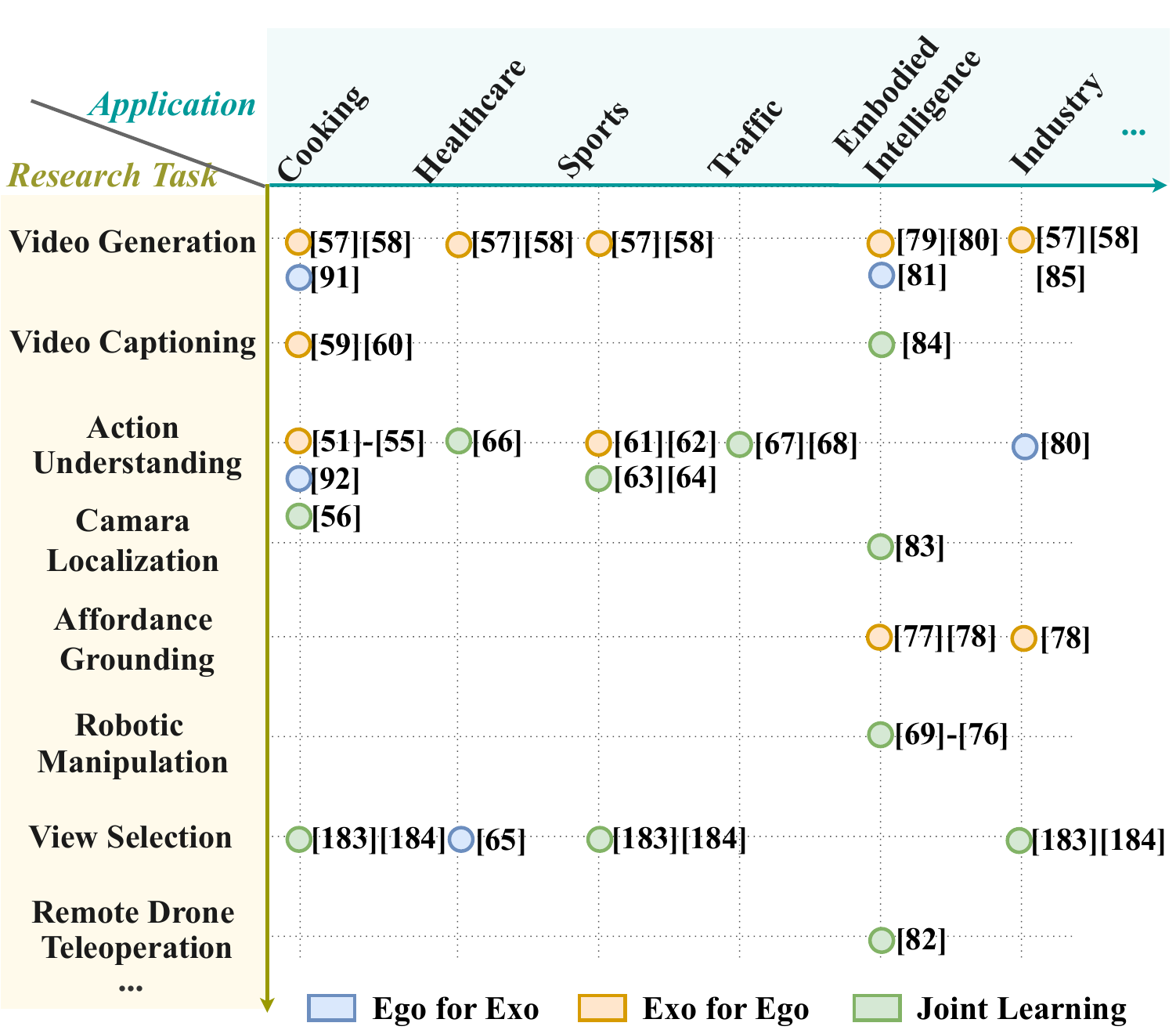}
\caption{Mapping relevant research works to applications and research tasks.}
\label{fig:mapping}
\vspace{-3mm}
\end{figure}

In summary, existing works demonstrate a promising foundation for exploring egocentric and exocentric collaboration in specific applications. However, despite this progress, current developments remain insufficient to meet the growing demands of real-world deployment. As illustrated in Fig.~\ref{fig:mapping}, many tasks critical to applications remain under-investigated. Consequently, the next section presents a detailed review of research tasks and their associated advancements, emphasizing the capabilities and limitations of existing research.

\section{Research Tasks} \label{section: task}

The previous section introduces research progress tailored to specific applications. Building on this foundation, this section provides a comprehensive review of advancements in cross-view collaboration with both egocentric and exocentric perspectives. We organize the research directions into three categories, which are defined as follows:
\begin{enumerate}
    \item \textbf{Exocentric for Egocentric:} This direction focuses on leveraging knowledge from the exocentric domain to enhance egocentric video understanding.
    \item \textbf{Egocentric for Exocentric:} Inversely, this direction emphasizes utilizing knowledge from the egocentric domain to improve exocentric video understanding.
    \item \textbf{Joint Learning:} This direction aims to integrate egocentric and exocentric perspectives to address cross-view video understanding tasks.
\end{enumerate}

\begin{figure*}[t]
\centering
\includegraphics[width=2.0\columnwidth]{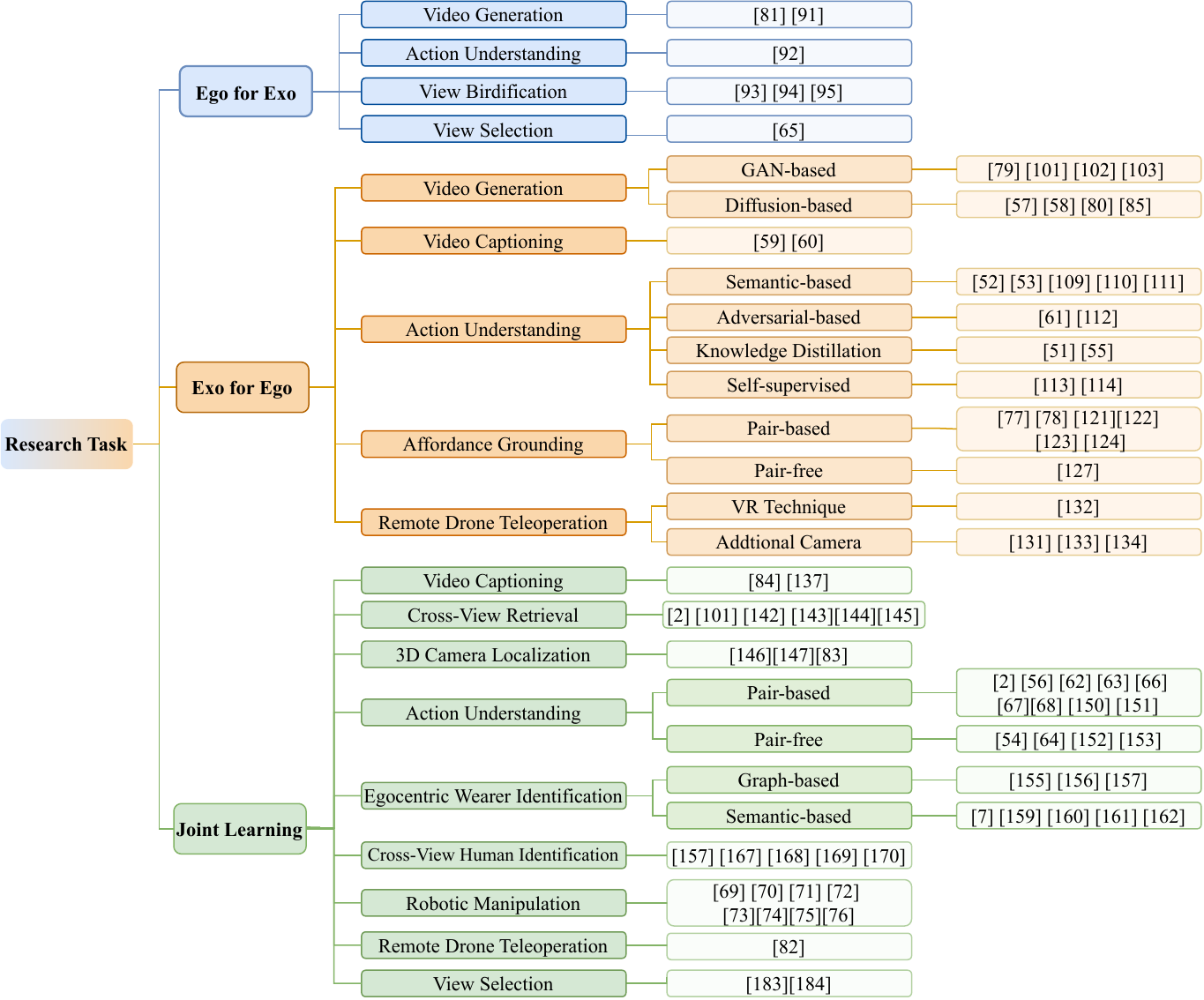}
\caption{Overall structure of Section \nameref{section: task}. We discuss the research task from three aspects: Egocentric for Exocentric, Exocentric for Egocentric, and Joint Learning. Each subsection reviews a variety of tasks and their existing works.}
\label{fig:category}
\vspace{-3mm}
\end{figure*}

For each direction, we cover various research tasks and review the existing work. An overview is illustrated in Fig.~\ref{fig:category}.

\subsection{Egocentric for Exocentric}

The unique viewpoints of egocentric videos provide rich details that are often missing from exocentric perspectives. This subsection reviews research efforts that leverage egocentric perspectives to enhance exocentric tasks.

\noindent\textbf{Video Generation.}
Ego-to-exo video generation involves generating an exocentric video from an egocentric one, offering a different perspective of the same environment. It offers significant research value across various fields. For instance, in virtual touring, travelers can review their routes from the third-person perspective to plan their trips effectively. 

Video generation has made significant progress in recent years~\cite{Hu2023AnimateAC, Blattmann2023StableVD, Yang2024CogVideoXTD}. However, ego-to-exo video generation poses unique challenges. Egocentric view often includes obscured regions, making it difficult to reconstruct the broader scene of the exocentric perspective. Additionally, maintaining consistency across views is challenging due to their significant disparity. Recent studies in video generation use depth maps \cite{gen_depth}, poses \cite{Hu2023AnimateAC}, and other conditional inputs \cite{Blattmann2023StableVD, gen_text} to provide spatial-temporal constraints. However, acquiring such cues in both egocentric and exocentric settings remains difficult.

\begin{figure}[H]
\centering
\includegraphics[width=1.0\columnwidth]{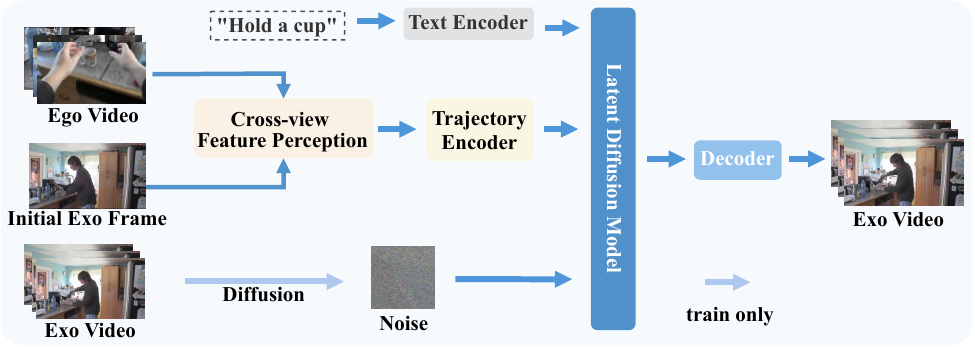}
\caption{Illustration of a diffusion-based framework for ego-to-exo video generation, adapted from \cite{ide}.}
\label{fig:ego4exo_gen}
\end{figure}

For ego-to-exo video generation, IDE \cite{ide} introduces a novel framework that leverages human intention to maintain consistency across perspectives. It proposes that human intention is view-independent and can be used to establish connections between views. Specifically, it represents human intention through human movement and action descriptions, which serve as conditional inputs for the diffusion model, as illustrated in Fig. \ref{fig:ego4exo_gen}.
Different from IDE \cite{ide}, another work \cite{water} investigates this task for underwater vehicles. Although onboard cameras provide a first-person view, this limited perspective restricts the operator's ability to maneuver in complex underwater environments. To address this, this approach uses past egocentric views and camera poses to create an eye-on-the-back view. This synthesis exocentric views provide broader scene context and enhance operational efficiency.

\noindent$\bullet~$Discussion: 
Despite prior efforts in video generation, ego-to-exo video generation remains under-explored, particularly in applications such as robotics and autonomous driving. In these domains, first-person videos (e.g., from onboard cameras) are the primary data source, but their limited view restricts comprehensive scene understanding. In contrast, third-person videos provide broader context, enabling better analysis and decision-making. 
To realize ego-to-exo video generation in real-world systems, future research must address domain-specific challenges. For example, resource-constrained edge devices cannot support state-of-the-art video generation models, necessitating the development of lightweight architectures. Furthermore, delayed inference in ego-to-exo synthesis could disrupt robotic control or vehicle safety. These challenges highlight the need for real-time processing in future solutions.

\noindent\textbf{Action Understanding.}
Human action analysis is widely studied with third-person data~\cite{kinetic,HowTo100M,ava,Soomro2012UCF101AD}. The exocentric perspective captures the full body movements but often misses action details. In contrast, egocentric videos excel at capturing detailed human-object and human-human interactions, which offer a complementary viewpoint to enhance exocentric action understanding. 

\begin{figure}[h]
\centering
\includegraphics[width=1.0\columnwidth]{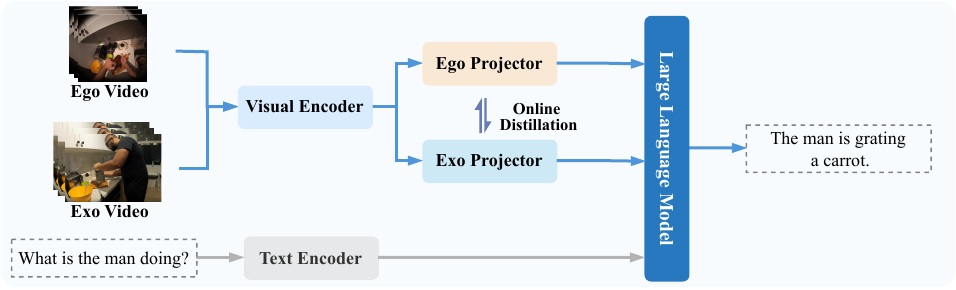}
\caption{Illustration of a typical method for ego-for-exo action understanding, adapted from \cite{reilly2025}. This method distills egocentric cues into exocentric representations.}
\label{fig:ego4exo_action}
\vspace{-5mm}
\end{figure}

To leverage complementary egocentric perspectives, Reilly et al. \cite{reilly2025} propose a distillation approach, as illustrated in Fig.~\ref{fig:ego4exo_action}. This approach employs projectors to align video features with large language models embeddings, followed by knowledge distillation to transfer egocentric cues into exocentric representations. It highlights the potential of egocentric cues in improving exocentric activity understanding for large vision-language models.

\noindent$\bullet~$Discussion: using egocentric perspectives to complement exocentric action analysis is under-explored in fields like industry and surgery. In these domains, performance evaluation is typically conducted via third-person cameras or in-person monitoring. However, the egocentric perspective can capture more fine-grained details from the actor's viewpoint. To enable ego-for-exo action analysis, future research should develop lightweight wearable devices that don't disrupt operations and address issues like motion blur and rapid viewpoint shifts in egocentric videos to improve alignment with exocentric views.

\noindent\textbf{View Birdification.} This task aims to estimate the trajectories of a crowd from a bird-eye’s view from egocentric videos captured by an observer. It recovers the global movements of people from the observations of the observer. This task has a wide range of applications such as crowd behavior analysis and surveillance.

\begin{figure}[h]
\centering
\includegraphics[width=1\columnwidth]{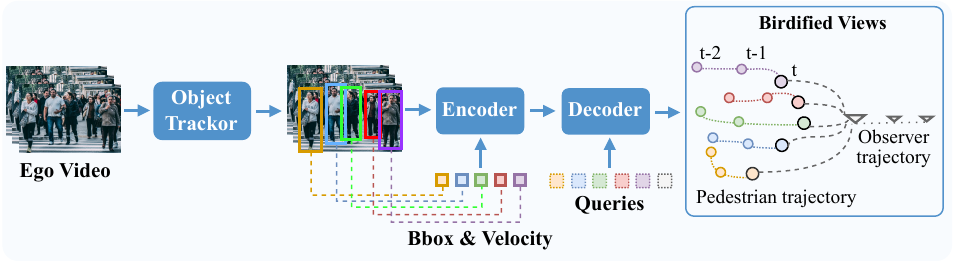}
\caption{Illustration of a typical method for view birdification, adapted from \cite{ViewBirdiformer}. This task aims to estimate the trajectories of a crowd in a bird's-eye view from an observer's egocentric perspective.}
\label{fig:ego4exo_bird}
\vspace{-4mm}
\end{figure}

In \cite{View_Birdification}, a cascaded optimization based method is proposed to alternate between estimating the displacements of the egocentric camera and its surrounding pedestrians. However, this iterative approach incurs high computational cost. 
To address this issue, ViewBirdiformer \cite{ViewBirdiformer} proposes a transformer-based architecture that performs view birdification in a single forward pass. As illustrated in Fig.~\ref{fig:ego4exo_bird}, it first utilizes a multi-object tracking algorithm to extract pedestrian movements, including bounding box coordinates and velocity vectors. These features are then encoded via a transformer encoder to model pedestrian interactions. Subsequently, the transformer decoder leverages camera queries and pedestrian trajectory queries from the previous timestep to predict pedestrian trajectories for the next timestep.
In subsequent work, InCrowdFormer \cite{Nishimura2023InCrowdFormerOP} addresses uncertainties caused by unknown pedestrian heights and simultaneously predicts pedestrian trajectories along with their associated uncertainty probabilities.

\noindent$\bullet~$Discussion: view birdification has promising applications in crowd management and security monitoring. These scenarios mainly rely on fixed surveillance cameras, which are often hindered by limited coverage. In contrast, mobile egocentric cameras can effectively capture blind spots and dynamically track targets.
To support on-site applications, future research must address the unique challenges inherent to egocentric videos. For instance, the mobile nature of egocentric cameras introduces issues such as rapid viewpoint changes and environmental transitions (\textit{e.g.}, indoor-to-outdoor shifts). These factors can degrade video quality and hinder trajectory estimation. Future work could integrate video enhancement techniques \cite{li2023fastllve, video_deblur, motion_blur} to mitigate these challenges.

\noindent\textbf{View Selection.} Surgery recordings serve as an essential resource for medical education and surgical assessment. To minimize occlusion and fully capture the surgical field, recording systems often employ multiple cameras mounted in the surgical lump. Therefore, a crucial task is to automatically select the optimal camera view at every moment.

\begin{figure}[h]
\centering
\includegraphics[width=1\columnwidth]{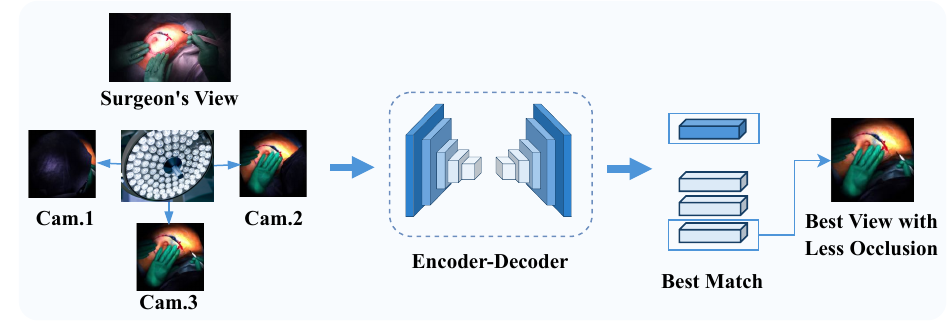}
\caption{Illustration of a typical method for view selection in surgical recording, adapted from \cite{Saito}. It aims to identify the exocentric view with minimal occlusion by using the surgeon's egocentric perspective as a selection criterion.}
\label{fig:ego4exo_selection}
\vspace{-4mm}
\end{figure}

As discussed in \cite{Saito}, the doctor’s perspective is considered the most effective to capture surgical targets.  Therefore, this method selects the exocentric camera view that best matches the doctor's egocentric perspective, as demonstrated in Fig.~\ref{fig:ego4exo_selection}. Future work can leverage sequential information from egocentric videos to reduce frequent camera switching and incorporate other learning algorithms\cite{DBLP:conf/visapp/ShimizuOHKTS20, Hachiuma2020DeepSA}.

\subsection{Exocentric for Egocentric}
Exocentric perspectives can complement egocentric analysis by providing a broader view of the environment. Additionally, large-scale exocentric video datasets~\cite{kinetic,HowTo100M,ava,Soomro2012UCF101AD} has driven significant progress in exocentric video understanding~\cite{Two_streamCNN,kinetic,vivit,videotransformer}. Building on these advancements, recent studies have investigated leveraging data and models from the exocentric domain to enhance egocentric analysis. This subsection reviews key approaches that utilize exocentric video techniques to improve egocentric tasks.

\noindent\textbf{Video Generation.}
Exo-to-ego generation aims to create a first-person view from third-person recordings. This task benefits various fields. For example, in VR and AR applications, exo-to-ego generation can help the users understand procedures by converting third-person videos into their own perspectives. Similarly, the embodied agents can leverage exo-to-ego generation to better understand their surrounding environment. 

\begin{figure}[b]
\centering
\includegraphics[width=1.0\columnwidth]{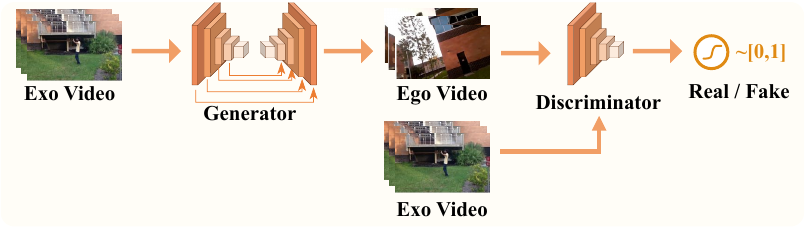}
\caption{Illustration of the general GAN-based framework for exo-to-ego video generation. The generator uses exocentric images to synthesize egocentric views, while the discriminator distinguishes between real and synthesized egocentric images.}
\label{fig:exo4ego_gen}
\end{figure}

Current exo-to-ego generation approaches can be categorized into \textit{GAN-based} \cite{Elfeki2018FromTP, liu2021cross, liu2022parallel,garello2022towards} and \textit{diffusion-based} \cite{luo2025put,Exo2Ego-V, Cheng20244DIFF3D, spisak2024diffusing,xu2025egoexo} methods.
In \cite{Elfeki2018FromTP, garello2022towards}, exocentric images are used as conditional inputs to GAN for synthesizing egocentric images. Fig.~\ref{fig:exo4ego_gen} illustrates the general framework of GAN-based approaches.
Liu et al. \cite{liu2022parallel} proposes a two-parallel-GANs architecture to transform images from one viewpoint to another. 
However, these works \cite{Elfeki2018FromTP, garello2022towards, liu2022parallel} are limited to image generation. For video generation, STA-GAN \cite{liu2021cross} proposes a bi-directional GAN to learn both spatial and temporal information. However, it relies on semantic maps for guidance to overcome generation ambiguities. 
More recent work \cite{luo2025put,Exo2Ego-V, Cheng20244DIFF3D,spisak2024diffusing} leverages diffusion models. Exo2Ego \cite{luo2025put} and Exo2Ego-V \cite{Exo2Ego-V} focus on synthesizing videos of human activities, while \cite{spisak2024diffusing} targets robot manipulation scenarios.

\noindent$\bullet~$Discussion: despite ongoing research efforts, transforming instructor demonstration videos into egocentric views for educational purposes remains under-explored in applications such as industrial training, engineering, and surgery. In these fields, egocentric perspectives can provide trainees with immersive experiences and enhance their understanding of complex procedures. 
First, viewpoint transformation from exocentric to egocentric is inherently ill-posed, as it requires synthesizing visual content that is not directly observed in the source view. This demands robust geometry-aware models capable of inferring occluded or unobserved regions while maintaining spatial coherence. Second, accurately modeling head motion and gaze dynamics is critical for generating realistic egocentric views, yet remains difficult due to the lack of ground truth head-pose trajectories in most instructional videos. Third, current systems struggle with fine-grained temporal alignment, making it difficult to synchronize key actions across views, especially in long, unstructured demonstrations. Finally, achieving semantic consistency—ensuring that important task-relevant elements (e.g., tools, hands, and object interactions) are preserved and emphasized in the transformed view—is an open challenge, particularly in cluttered or multi-agent scenes.

\noindent\textbf{Video Captioning.}
This task involves generating descriptive textual narratives for videos, aiming to produce coherent sentences that describe the actions, objects, and interactions in the video.

Traditionally, video captioning has been extensively studied in the context of third-person videos \cite{caption_1, caption_2, caption_3}, supported by large-scale exocentric video datasets. In contrast, egocentric video captioning has received less attention due to the limited availability of large-scale, high-quality egocentric datasets.

\begin{figure}[b]
\centering
\includegraphics[width=1.0\columnwidth]{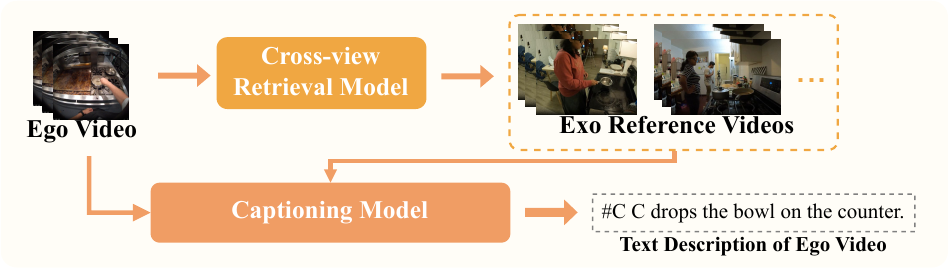}
\caption{Illustration of a typical method for exocentric for egocentric video captioning, adapted from \cite{xu2024retrieval}. This method retrieves relevant exocentric videos to serve as references for captioning egocentric videos.}
\label{fig:exo4ego_caption}
\end{figure}

Currently, a promising direction for egocentric video captioning is leveraging large-scale third-person data.
To mitigate domain shift, Ohkawa et al. \cite{ohkawa2023exo2egodvc} introduce an intermediate ego-like view to gradually adapt from exocentric to egocentric views.
On the other hand, EgoInstructor \cite{xu2024retrieval} is a retrieval-augmented captioning model that uses semantically relevant exocentric videos as references for egocentric video captioning, as shown in Fig. \ref{fig:exo4ego_caption}.

\noindent$\bullet~$Discussion: Egocentric video captioning has significant potential for assistive devices designed to enhance environmental awareness for visually impaired individuals. In such scenarios, wearable devices, such as smart glasses, can use egocentric video feeds to generate real-time descriptions of user's surroundings \cite{worldscribe, wanderguide}.  
However, as discussed in \cite{worldscribe, wanderguide}, the limited field of view of egocentric cameras primarily captures salient foreground objects while often fails to capture broader scene layouts. This constraint impairs users' ability to reconstruct spatial relationships. A promising approach to addressing this limitation is augmenting egocentric captioning with exocentric 3D spatial data. However, integrating exocentric data into assistive systems should address challenges like translating exocentric 3D layouts into user-centric spatial references (e.g., egocentric distance and orientation) to meet user-specific demands.

\noindent\textbf{Action Understanding.}
Due to the availability of large-scale exocentric datasets \cite{kinetic, HowTo100M,ava,Soomro2012UCF101AD}, exocentric action understanding has been extensively studied. Consequently, a body of research explores leveraging knowledge from the exocentric domain to improve understanding of egocentric action.

\textit{Semantic-based methods} focus on leveraging shared semantics between egocentric and exocentric videos to bridge the gap between the two domains.
Existing studies have explored the use of activity sounds~\cite{zhang2022audio}, geometric correlations \cite{truong2023cross}, skeleton poses \cite{rocha2023cross}, and narrations \cite{sum-l} to establish relationships between egocentric and exocentric perspectives. In addition, EMBED \cite{dou2024unlocking} utilizes hand-object interactions to transform exocentric video-language datasets into egocentric style.

\begin{figure}[b]
\centering
\includegraphics[width=1.0\columnwidth]{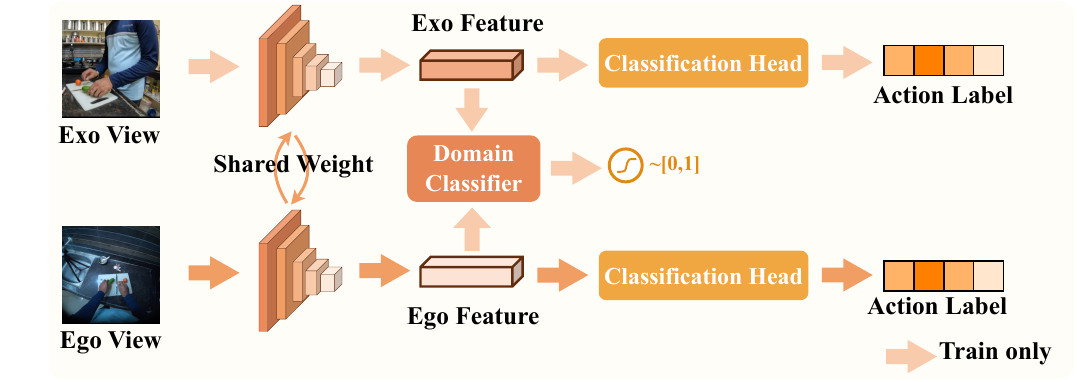}
\caption{Illustration of a general adversarial-based approach for exo-for-ego action understanding. During training, the domain classifier differentiates between egocentric and exocentric features, while the model is optimized to deceive it. During inference, only egocentric videos are used.}
\label{fig:exo4ego_action}
\end{figure}

\textit{Adversarial-based methods} employ adversarial strategies to minimize the discrepancy between the exocentric domain (source domain) and the egocentric domain (target domain). In \cite{choi2020unsupervised, wang2022estimating}, a domain classifier is utilized to differentiate whether the feature originates from egocentric or exocentric videos. During training, the model is optimized to generate features to fool the domain classifier, thereby aligning egocentric features with exocentric features. Fig.~\ref{fig:exo4ego_action} demonstrates the general adversarial-based framework.

\textit{Knowledge distillation methods} seek to distill knowledge from exocentric models to improve egocentric action understanding. In \cite{quattrocchi2023synchronization, li2021ego}, the model is first trained on exocentric videos. Subsequently, knowledge distillation losses are applied to adapt the model for egocentric videos.

\textit{Self-supervised methods} address the challenge of requiring large-scale labeled egocentric data. Egofish3D \cite{liu2023egofish3d} utilizes 3D poses estimated by an exocentric pose estimator as supervision signals to train an egocentric pose estimator without 3D ground truth annotations. Ex2Eg-MAE \cite{tran2024ex2eg} first learns to reconstruct exocentric frontal facial videos using synthesized multi-view data that emulate egocentric environments and then evaluates on egocentric social role understanding tasks.

\noindent$\bullet~$Discussion: 
existing exo-for-ego frameworks mainly focus on basic tasks such as action recognition \cite{jia2024audio, rocha2023cross, sum-l, wang2022estimating} and pose estimation \cite{liu2023egofish3d}.
However, with the growing demand for advanced applications like skill assessment \cite{skate} and automated commentary generation \cite{expertaf, rao2024unisoccer}, we propose expanding the use of exocentric data to tackle more complex challenges. For instance, exocentric expert demonstrations could serve as references to guide egocentric actions and deliver tailored feedback. To advance this in real-world systems, future research should establish skill-level evaluation criteria and improve cross-view action alignment.

\noindent\textbf{Affordance Grounding.}
This task aims to identify and localize the interaction regions of objects based on given instructions. In this task, the exocentric view captures the interactions between human and object while the egocentric view refers to the object only images. Affordance grounding plays a critical role in applications such as embodied intelligence~\cite{6doffinegrainedgrasp, Robo-ABC}, where robots must not only recognize objects but also understand how to interact with them. 

\begin{figure}[b]
\centering
\includegraphics[width=1.0\columnwidth]{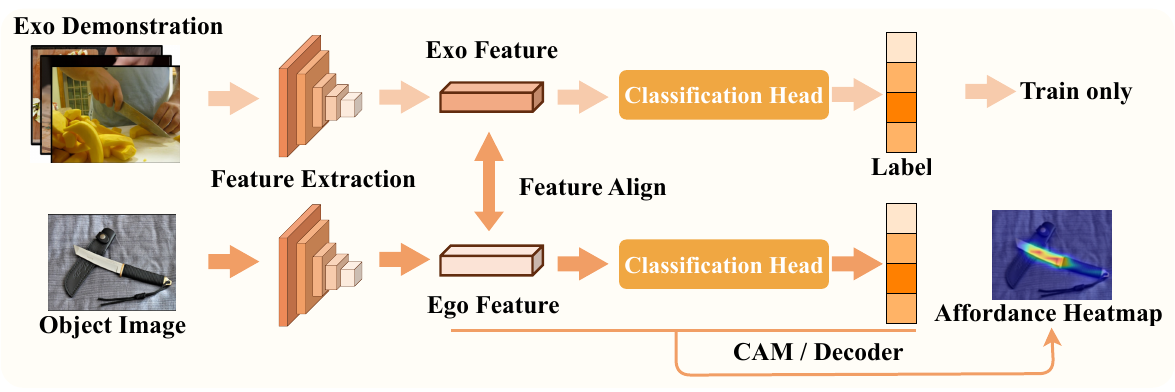}
\caption{Illustration of the exo-for-ego affordance grounding framework. During training, egocentric object images are aligned with exocentric demonstration images with the same affordance label. During inference, only the egocentric image is inputted to identify the affordance region.}
\label{fig:exo4ego_afford}
\end{figure}

Exo-for-Ego affordance grounding methods can be categorized into two types based on training data: \textit{pair-based} and \textit{pair-free}. Fig. \ref{fig:exo4ego_afford} presents a general framework for this task. 

\textit{Pair-based method} \cite{Luo_2022_CVPR, Li_2023_CVPR, yang2024learning,xu2024weakly,zhang2024self,Rai_2024_CVPR} learn from a group of exocentric images and the corresponding egocentric object image that share the same affordance label. During inference, only the egocentric object image is used. 
Luo et al. \cite{Luo_2022_CVPR} introduce  Cross-View-AG based on Class Activation Mapping (CAM) \cite{zhou2016learning}, which has served as a foundational paradigm for many subsequent studies. 
However, CAM is only used in post-processing during inference and lacks effective supervision for the generated affordance map.
To address this, LOCATE \cite{Li_2023_CVPR} replaces the vanilla CAM with a learnable module to enable supervision of the CAM-generated map. 
Furthermore, GAAF-Dex \cite{yang2024learning} enhances \cite{Li_2023_CVPR} by applying concentration loss to make the affordance map more compact.

With advances in large language models (LLMs), a number of studies \cite{xu2024weakly, zhang2024self,Rai_2024_CVPR} integrate language signals into affordance grounding learning. 
WSMA uses CLIP~\cite{clip} to encode affordance labels and fuses them with egocentric image embeddings.  However, it does not address the issue of action ambiguity, where an object may support multiple actions. To address this limitation, Zhang et al. \cite{zhang2024self} enable the model to predict both affordance region and object-action descriptions. In contrast to \cite{zhang2024self}, Rai et al. \cite{Rai_2024_CVPR} utilize world knowledge from LLMs to generate more detailed captions that include information about object parts and attributes. 

Unlike previous work, \textit{pair-free methods} do not require paired inputs. Instead, they learn from a group of exocentric images and their affordance information. INTRA \cite{jang2024intra} uses contrastive learning as a weakly supervised objective to extract shared knowledge from different affordance labels. 
\noindent$\bullet~$Discussion: using human demonstration videos to learn robot-centric affordance has not been fully investigated in scenarios like industrial automation. In these domains, egocentric videos from onboard robot cameras are the primary data source. However, human operation videos can guide robots in mastering precise tasks, such as assembly and material handling. 
While promising, existing research has yet to address the domain-specific challenges. For example, operating precision instruments demands high affordance accuracy, as even minor deviations can lead to operational failures. Due to factors like cross-view object scale discrepancy~\cite{afford_survey}, current methods struggle to effectively transfer affordance regions across views to achieve such precision requirement.

\noindent\textbf{Remote Drone Teleoperation.}
Drones can navigate challenging environments or locations impassable for humans. It has a wide range of applications such as disaster investigations\cite{EmergencyNet} and product delivery\cite{product}. Typically, drone control systems offer an egocentric view through an on-board camera. However, this limited field of view fails to fully capture the surroundings.

\begin{figure}[b]
\centering
\includegraphics[width=1.0\columnwidth]{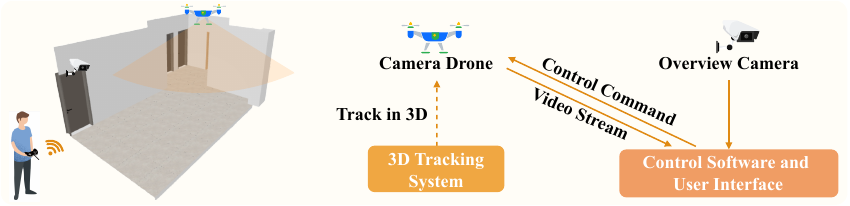}
\caption{Illustration of remote drone teleoperation with additional cameras, adapted from \cite{StarHopper}. To overcome the limited view of egocentric cameras on drones, exocentric cameras are used to capture the surrounding environment.}
\label{fig:exo4ego_drone}
\end{figure}

To address the limitations of egocentric views, previous research has explored using \textit{VR technique} \cite{drone_arvr} or \textit{additional cameras} \cite{StarHopper,third_pilot,birdviewar} to provide exocentric views. Fig.~\ref{fig:exo4ego_drone} illustrates using overhead camera to provide exocentric views for drone teleoperation.
In \cite{drone_arvr}, VR technique provides a 3D model of the environment, allowing pilots to perceive the drone's surroundings. Another line of works \cite{StarHopper,third_pilot,birdviewar} utilize additional cameras to capture the environment of the drone. StarHopper\cite{StarHopper} uses a fixed overhead camera while Temma et al.\cite{third_pilot} uses a secondary drone that semi-automatically flies around the primary drone. Inspired by \cite{third_pilot}, BirdViewAR\cite{birdviewar} further uses AR overlays to highlight the primary drone’s spatial status and proposes an automatic framing method to ensure the secondary drone follows the primary drone in fast-moving scenarios.

\subsection{Joint Learning}
Joint learning aims to leverage both egocentric and exocentric perspectives to address cross-view video understanding tasks. It requires both egocentric and exocentric views as input during both training and inference. This contrasts with unidirectional paradigms (e.g., exo-for-ego or ego-for-exo), where often one view serves as auxiliary information during training, but only a single view is utilized at test time. In joint learning, however, it emphasizes bidirectional collaboration to resolve cross-view tasks. Below, we systematically review advancements in cross-view tasks, highlighting diverse strategies for effectively integrating the complementary nature of egocentric and exocentric perspectives.

\noindent\textbf{Video Captioning.} In daily life, video captioning can document a wide range of human activities in natural language. This capability can enhance the development of smart assistants \cite{vinci, LITA, huang2025egocentric} to help humans memorize and retrieve items.

\begin{figure}[h]
\centering
\includegraphics[width=1.0\columnwidth]{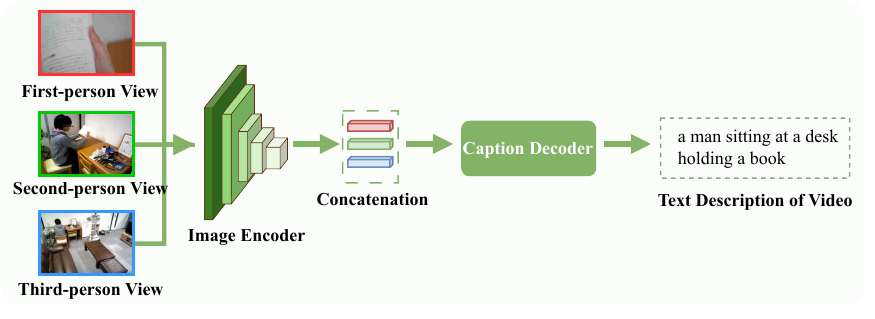}
\caption{Illustration of video captioning using videos from first-person, second-person, and third-person perspectives.}
\label{fig:joint_caption}
\vspace{-2mm}
\end{figure}

Current research \cite{lifelog, Nakashima2020LifeloggingCG} investigates captioning lifelog videos in multi-view settings. The logging system comprises a first-person view from an individual, a second-person view from a service robot, and a third-person view from a fixed camera, as demonstrated in Fig. \ref{fig:joint_caption}. 
In \cite{lifelog}, multi-view images are independently processed into image features, which are then concatenated and projected into a unified feature space. The unified features are subsequently input into a caption decoder to generate captions. 
In contrast, Nakashima et al. \cite{Nakashima2020LifeloggingCG} employ attention mechanisms for feature fusion. This method first uses Faster R-CNN \cite{FasterRCNN} to detect salient regions from each view. To address redundant cross-view information, the detected features are clustered into several groups and then fused via attention mechanisms. 

\noindent$\bullet~$Discussion:
For ego-exo video captioning, several challenges remain for future research. One key issue is balancing description granularity. Due to the different fields of view, egocentric and exocentric videos may emphasize different visual elements. This requires models to reconcile these disparities to generate consistent captions. Additionally, as discussed in \cite{worldscribe, wanderguide}, users may prefer different levels of detail. Future research should enable model to adjust description granularity to align with user-specific needs.
Another challenge is managing redundant and complementary information across views. While prior work \cite{Nakashima2020LifeloggingCG} addresses this by clustering features at frame-level, it overlooks action-level correspondences. For example, an egocentric view might depict ``hand pulls a lever", while an exocentric view captures ``doors open". To generate coherent captions, models must integrate cross-view action dependencies. To achieve this, future work can integrate techniques like action segmentation \cite{quattrocchi2023synchronization, Sarfraz2021,Huang_2020_improving} and action relation \cite{VCL, SVIP,huang2020mutual}. Beyond technical challenges, joint video captioning holds significant promise for smart assistants \cite{vinci, LITA}. By integrating multiple perspectives, such systems can generate comprehensive activity logs, enabling assistants to memorize historical events and support downstream tasks like temporal grounding and visual question answering.

\begin{figure}[h]
\centering
\includegraphics[width=1.0\columnwidth]{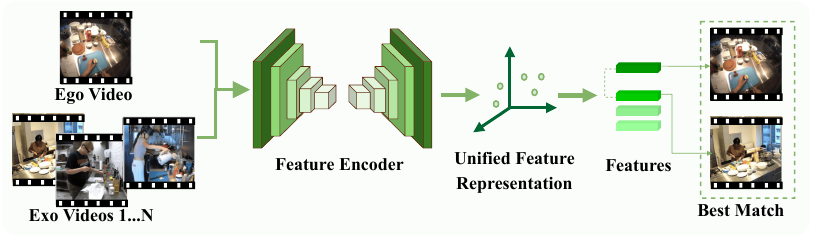}
\caption{Illustration of the general cross-view retrieval framework. Exocentric and egocentric videos are encoded into a shared representation space to retrieve the best match from the alternate view.}
\label{fig:joint_retreival}
\vspace{-2mm}
\end{figure}

\noindent\textbf{Cross-View Retrieval.} 
This task focuses on identifying and retrieving corresponding visual elements, such as videos\cite{Ardeshir2016EgoTransferTM}, frames\cite{Elfeki2018FromTP, actorobserver,tjanet}, and moments\cite{actorobserver,tjanet}, from different viewpoints, as demonstrated in Fig. \ref{fig:joint_retreival}. 

Early work \cite{Ardeshir2016EgoTransferTM} explores linear and non-linear mappings to transform motion features between two views. More recent approaches \cite{Elfeki2018FromTP, actorobserver} first utilize separate branches to extract features from different views and then employ contrastive learning to align representations. Furthermore, T-JANet \cite{tjanet} leverages overlapping attention regions between views to guide representation learning. 
However, these works mainly address cross-view correspondence at the video level. Recently, Ego-Exo4D \cite{Egoexo4d} introduces a cross-view object correspondence task, which aims to predict object masks in one view given queries from another view. 
PSALM \cite{Psalm} demonstrates zero-shot capability for this task. It first utilizes LLM to process visual and textual prompts, followed by a general segmentation model to generate object masks. Building on PSALM \cite{Psalm}, ObjectRelator \cite{ObjectRelator} generates descriptive language prompts for query objects to exploit the LLM’s reasoning ability. To address object appearance disparities across views, ObjectRelator \cite{ObjectRelator} further introduces a cross-view object alignment module to project masks from different views into unified space.

\noindent$\bullet~$Discussion: 
Current approaches primarily learn shared representations across views. However, inherent view disparities lead to significant differences in appearance and motion. These challenges are further exacerbated by occlusions and out-of-view scenarios. Such issues complicate representation alignment. Future work could explore disentangling features into view-invariant and view-specific components \cite{holographic}.
Beyond technical challenges, cross-view retrieval is under-explored in applications like surveillance systems. For instance, retrieving relevant surveillance clips based on egocentric videos from law enforcement agents could enhance event understanding, crime localization, and object tracking. However, retrieving from large-scale data is computationally intensive. Future work should optimize retrieval speed for practical deployment.

\noindent\textbf{3D Camera Localization.} 
This task aims to determine the position and orientation of a camera in the environment.  

\begin{figure}[h]
\centering
\includegraphics[width=1\columnwidth]{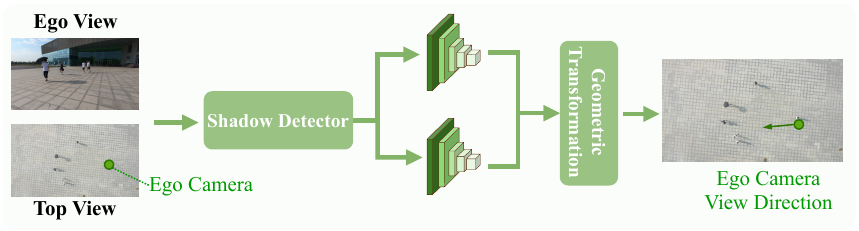}
\caption{Illustration of a typical method for egocentric camera localization, adapted from \cite{bev1}. This method uses shadows to relate egocentric and top views, and estimates the egocentric camera direction in the top view.}
\label{fig:ego4exo_cam}
\vspace{-2mm}
\end{figure}

Han et al. \cite{bev1} and Qian et al. \cite{bev2} propose to localize egocentric cameras from a global top-down view. Han et al. \cite{bev1} leverage shadow to relate egocentric and top views and propose a shadow detection model to predict shadow direction, as shown in Fig.~\ref{fig:ego4exo_cam}. 
Furthermore, Qian et al. \cite{bev2} utilize the spatial distribution of subjects in the 3D environment to estimate egocentric camera poses in a virtual top-down view.
In contrast to \cite{bev1, bev2}, YOWO \cite{YOWO} introduces a novel approach to localize ceiling-mounted cameras (CMCs). Previous methods \cite{slam_1, EgoLocate} typically use SLAM for scene reconstruction and subsequently employ visual localization to estimate camera poses. However, the perspective disparity between egocentric and exocentric views poses challenge for cross-view localization. Moreover, the static nature of CMCs prevents using motion information to correct localization errors. 
To address these limitations, YOWO jointly optimizes scene reconstruction and CMC registration. It employs a mobile agent to navigates the environment to generate both agent trajectories and scene layout. Meanwhile, CMCs capture the agent to provide pseudo trajectories. By correlating these trajectories, YOWO aligns CMC poses with the scene layout.

\begin{figure}[h]
\centering
\includegraphics[width=1.0\columnwidth]{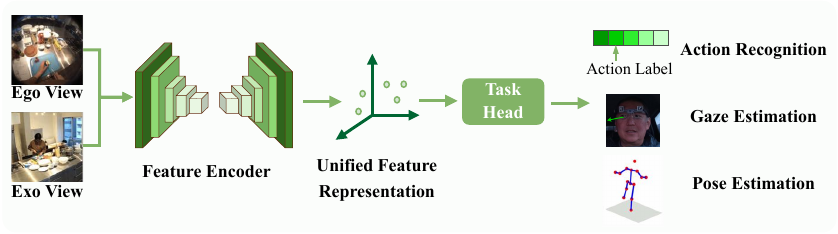}
\caption{Illustration of cross-view action understanding. This involves action recognition, gaze estimation, and pose estimation tasks.}
\label{fig:joint_action}
\vspace{-2mm}
\end{figure}

\noindent\textbf{Action Understanding.}
As discussed in \cite{egovlp, egovlpv2}, models predominantly trained on exocentric videos exhibit poor performance in egocentric data. Cross-view action understanding has emerged as a promising approach to enable a single model to achieve viewpoint-invariant action analysis. This field encompasses multiple key tasks, including action recognition, gaze estimation, and pose estimation, as illustrated in Fig. \ref{fig:joint_action}. Current research in this area can be broadly classified into two categories based on training data: \textit{pair-based} and \textit{pair-free}.

\textit{Paired-based methods} use synchronized egocentric and exocentric video pairs. For action recognition task, \cite{2014ar, anegolook,actorobserver, Yonetani} leverage paired videos to learn a unified feature across different views.
Soran et al. \cite{2014ar} jointly predict action labels and assess each camera's importance. 
In \cite{anegolook, actorobserver}, egocentric and exocentric videos are encoded by separate branches and subsequently aligned into a unified feature space.
Yonetani et al. \cite{Yonetani} use a pair of egocentric videos from two individuals to recognize micro-actions and reactions.
In driving scenarios, LBW \cite{lbw} utilizes both the driver's face and the forward road scene for gaze estimation. Similarly, Yang et al. \cite{yang2023aide} integrate in-vehicle and out-vehicle views to recognize the driver's state.
In the field of pose estimation, Ameya et al. \cite{dhamanaskar2023enhancing} map multi-view RGB frames and optical flow into a joint embedding space, while Hein et al. \cite{hein2023nextgeneration} evaluate multi-view methods \cite{10161514} for the pose estimation of surgical instruments.

While effective, paired-based approaches are limited by the expense of obtaining synchronized paired data. To address this limitation, recent research \cite{AE2,LaGTran, POV,holographic} has shifted towards leveraging unpaired videos.

\textit{Paired-free approaches} aim to learn shared action representations from unpaired egocentric and exocentric video data. During inference, pair-free models demonstrate flexibility by accepting either egocentric or exocentric video inputs for action analysis. This line of work can easily utilize existing large-scale third-person and first-person datasets. 
To align unpaired data, AE2 \cite{AE2} introduces a temporal alignment strategy. Based on the assumption that aligning egocentric and exocentric videos is inherently easier than aligning them when one sequence is temporally reversed, this approach employs reversed frames as negative samples for contrastive learning.
In contrast to AE2 \cite{AE2}, LaGTran \cite{LaGTran} leverages language descriptions to mitigate the domain gap between egocentric and exocentric videos. The method is based on the premise that text descriptions exhibit a smaller domain discrepancy compared to the original videos.
POV\cite{POV} incorporates learnable prompts to video tokens to learn view-agnostic representations. Unlike previous work, Huang et al. \cite{holographic} highlight the importance of view-specific information and disentangle features into view-invariant and view-specific components.  

\noindent$\bullet~$Discussion: 
Current methods \cite{2014ar, anegolook,actorobserver, Yonetani, lbw, yang2023aide, dhamanaskar2023enhancing, hein2023nextgeneration} mainly use paired videos to learn view-invariant representations. However, paired videos still exhibit large discrepancies due to perspective differences. Egocentric videos often suffer from blurring, distortion, and partial visibility, while exocentric videos may depict performers occupying minimal screen space, limiting fine-grained detail capture. These issues hinder shared representation learning. To bridge the disparity, promising solutions include video deblurring \cite{video_deblur, motion_blur} for egocentric videos, cropping action performers in exocentric videos \cite{dou2024unlocking} , and integrating IMU data \cite{zhang2025masked} to enhance motion information.
Furthermore, current research is confined to fundamental tasks like action recognition and pose estimation. Advanced tasks such as action assessment and feedback generation remain unexplored despite their potential in domains like sports. In this domain, integrating both perspectives can offer a holistic understanding of action regularity and proficiency, enabling personalized guidance. To enable practical deployment, a key challenge is effectively integrating dynamic granularity action information across views. Future work should balance between fine-grained hand-object interactions and full-body kinematics to achieve holistic analysis.

\noindent\textbf{Egocentric Wearer Identification.} Given both third-person and first-person videos captured in the same environment, this task aims to identify the egocentric camera wearer in third-person videos. It is similar to person re-identification across different views, but is more challenging since the camera wearer seldom appears in the egocentric view. 

\begin{figure}[b]
\centering
\includegraphics[width=1.0\columnwidth]{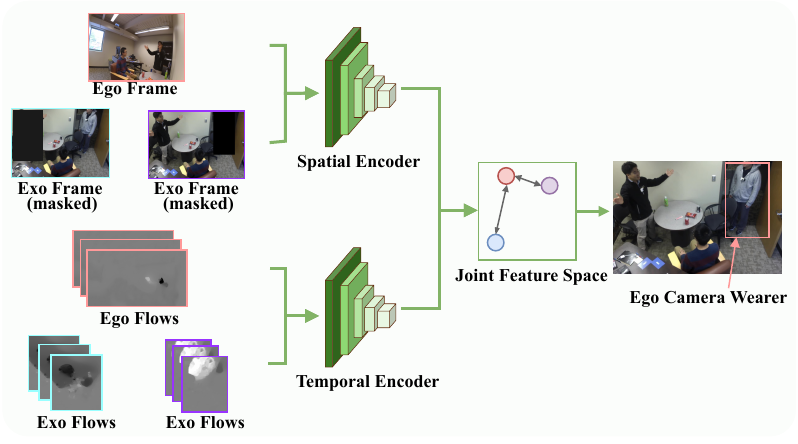}
\caption{Illustration of a typical method for egocentric wearer identification, adapted from \cite{Fan_2017_CVPR}. This method uses spatial and temporal information to learn view-invariant features and identify the egocentric wearer in exocentric images.}
\label{fig:joint_wearer}
\end{figure}

Early researches \cite{Ego2top, ego_meet, ego2top2} employ graph-based techniques to identify the camera holder of egocentric videos in top-view videos. \cite{Ego2top} models each video view as a graph and proposes a spectral graph matching technique. Building on this, \cite{ego_meet} extends the work of \cite{Ego2top} by considering time delays across videos. Furthermore, \cite{ego2top2} employs visual, geometric, and spatiotemporal reasoning to generate candidates and then uses graph cuts \cite{graphcuts} to evaluate candidates.

More recent approaches \cite{Fan_2017_CVPR, Joint_Seg,Joint_Seg2, yang2019visual, wen2021seeing} leverage shared semantic across views. Fan et al. \cite{Fan_2017_CVPR} leverage spatial (RGB frames) and temporal (optical flow) similarities to relate two views, as shown in Fig. \ref{fig:joint_wearer}. It employs contrastive learning to predict the camera wearer, utilizing first-person videos paired with third-person videos (masking the correct wearer) as positive samples, and third-person videos (masking a random person) as negative samples.
However, this approach primarily focuses on appearance similarity across views, overlooking the dynamic nature of the environment. To address this limitation, Visual-GPS \cite{yang2019visual} leverages motion and action information to improve robustness, as these features are less sensitive to environmental variations.
Subsequent work \cite{wen2021seeing} proposes a more challenging setting: predicting the camera wearer's location and pose in a third-person scene frame, where the wearer is absent.
Furthermore, \cite{Joint_Seg} and \cite{Joint_Seg2} jointly address person identification and segmentation and prove that solving these two problems simultaneously is mutually beneficial. 

\noindent$\bullet~$Discussion:  
Current appearance-based methods \cite{Fan_2017_CVPR, Joint_Seg, Joint_Seg2} may fail when the wearer is partially visible in the exocentric view. In such cases, even motion cues may struggle if critical body parts are occluded. Furthermore, in crowded scenarios, similar appearances (e.g., shared clothing) or similar actions (e.g., group sports) further hinder discriminative feature extraction. To address these limitations, future work could incorporate additional cues, such as object interactions \cite{egopca, shiota2024egocentric, egochoir} or person-person interactions \cite{av_conv, Yonetani}, to provide more distinctive information.
Beyond technical challenges, egocentric wearer identification remains unexplored in applications like rescue and emergency. In these fields, when critical events are detected in egocentric videos, command centers can locate the wearer in third-person views to dispatch assistance. To enable real-world deployment, future research must address domain-specific challenges. For instance, in large-scale emergencies, systems must distinguish between multiple egocentric wearers in overlapping exocentric views, requiring multi-agent identification algorithms. Additionally, the system must process high-volume, streaming data with minimal latency for quick response, necessitating online processing approaches.

\begin{figure}[t]
\centering
\includegraphics[width=1.0\columnwidth]{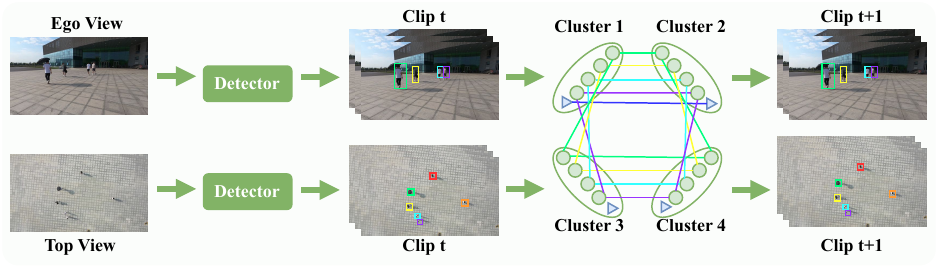}
\caption{Illustration of a typical method for cross-view human tracking and association, adapted from \cite{Complementary_han}. This method segments video pairs into clips and tracks individuals across clips and views.}
\label{fig:joint_association}
\vspace{-4mm}
\end{figure}

\noindent\textbf{Cross-View Human Identification.} This task aims to detect and identify the same individuals across views. 
Current approaches \cite{ego2top2, CIP,Complementary_han,Complementary_han_2,connect_han} study this task on top-view and side-view. The top view, captured by drones at high altitudes, covers large areas and displays human spatial distribution. In contrast, side views from mounted cameras provide more details. 
Ardeshir et al. \cite{ego2top2} propose a graph-based technique while Han et al. \cite{CIP} use a multi-view human association algorithm to match individuals across different views. 
However, these works \cite{ego2top2, CIP} are limited to human identification across views and do not address tracking. Han et al. \cite{Complementary_han} propose a joint optimization model for identifying and tracking. This approach first segments video pairs into clips and tracks individuals across clips and views, as demonstrated in Fig. \ref{fig:joint_association}.
Additionally, \cite{Complementary_han_2} extends this work by incorporating spatial distribution for cross-view association and introducing a new approach for appearance reasoning. 
Previous approaches \cite{CIP, Complementary_han, Complementary_han_2} rely on offline detection models \cite{yolov3} to detect human bounding boxes, which may hinder association performance. To address this, Han et al. \cite{connect_han} propose a joint method for cross-view multi-human detection and association.

\noindent\textbf{Robotic Manipulation.}
This task involves controlling robots to interact with objects and perform actions, such as grasping or moving, to achieve specific goals. 

Multi-view robot manipulation has been widely studied. However, most approaches simply concatenate multi-view observations at the image level \cite{CoDER} or feature level \cite{robot_1, rt1, Zhao-RSS-23, RoboAgent, diffusionpolicy, ALOHAUA2}, without fully exploiting their complementary characteristics. We focus on approaches that explore integrating the complementary strengths of different perspectives.

\begin{figure}[t]
\centering
\includegraphics[width=1.0\columnwidth]{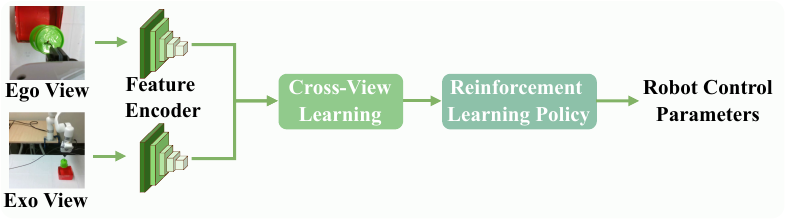}
\caption{Illustration of a typical framework of multi-view robotic manipulation, adapted from \cite{Jangir}. Multi-view data is integrated via cross-view learning module.}
\label{fig:joint_robot}
\vspace{-6mm}
\end{figure}

Lookcloser \cite{Jangir} utilizes cross-view attention mechanisms to integrate egocentric and exocentric perspectives, as shown in Fig.~\ref{fig:joint_robot}. In \cite{Hsu2022VisionBasedMN}, a variational information bottleneck is applied to third-person representations to mitigate their impact on out-of-distribution generalization.
Acar et al. \cite{Acar} utilize multi-view data to train a teacher policy, which then guides a single-view student policy through knowledge distillation. 
Sharma et al. \cite{Sharma2019} first use third-person human demonstration videos to generate task goal in robot's perspective, which are then combined with robot's current observation to predict actions.
Similarly, Shang et al. \cite{shang} leverage synchronized first-person and third-person demonstrations to learn viewpoint-agnostic representations and then use third-person demonstrations for policy learning. 
Both MV-MWM \cite{MV-MWM} and MFSC \cite{MFSC} introduce multi-view masked reconstruction strategies to learn representations from multi-view observations.
Unlike previous approaches, MVD \cite{dunion2024multiview} introduces a robust method that supports varying numbers of cameras in inference.

\noindent\textbf{Remote Drone Teleoperation.} Traditional drone manipulation primarily focuses on unidirectional collaboration, where humans send commands to control drones. In contrast, joint learning emphasizes bidirectional information exchange, allowing drones to access the human’s perspective for decision-making. This enhanced interaction supports a wider range of collaborative tasks. For instance, in a rescue mission, if a human operator identifies a potential victim through a wearable camera, the drone can autonomously navigate to the location to provide assistance. Such bidirectional communication improves operational efficiency.

\begin{figure}[h]
\centering
\includegraphics[width=1.0\columnwidth]{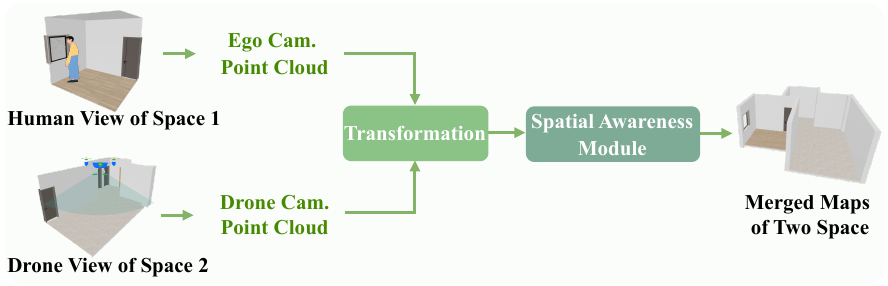}
\caption{Illustration of a typical method for remote drone teleoperation with human-drone collaboration, adapted from \cite{Morando}. This method combines user and drone perspectives into a unified environmental representation.}
\label{fig:joint_drone}
\end{figure}

A notable work in this field is presented in \cite{Morando}. In this study, point cloud data from the drone and the user’s wearable device are merged into a unified environmental representation, as demonstrated in Fig. \ref{fig:joint_drone}. Then, this approach provides visualizations of the environment from both the user’s and the drone’s perspectives, ensuring mutual awareness of the surroundings between the user and the drone.

\noindent$\bullet~$Discussion: To facilitate bidirectional information exchange between humans and drones, future research should optimize real-time data processing and minimize communication latency. Additionally, enhancing autonomous decision-making in drones based on both human and drone perspectives could further advance collaboration.

\noindent\textbf{View Selection.} 
The task of selecting the optimal viewpoint from multi-view videos has been widely studied. Prior work explores determining the best camera angles and positions in panoramic 360\textdegree~views \cite{hu2017deep, su2017making}, and automating viewpoint selection in multi-view systems \cite{yus2015multicamba, chen2019learning}. However, these methods typically address egocentric or exocentric views separately, ignoring scenarios where both views are available. 

\begin{figure}[h]
\centering
\includegraphics[width=1.0\columnwidth]{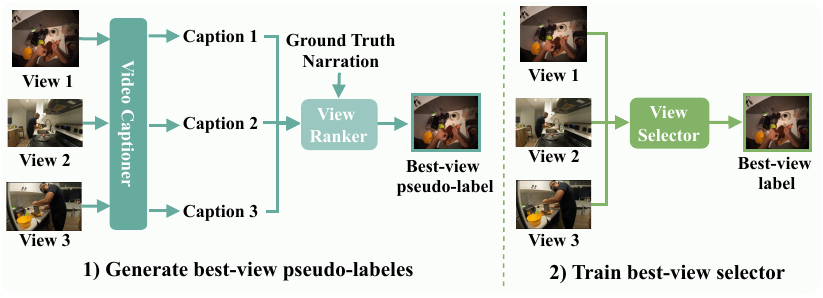}
\caption{Illustration of a typical ego-exo view selection method, adapted from \cite{majumder2024viewpoint}. This approach leverages video captions as weak supervision for selecting the best view.}
\label{fig:joint_sel}
\vspace{-3mm}
\end{figure}

Unlike previous work, recent work \cite{majumder2024viewpoint, switch-a-View} propose to address view selection in instructional videos, which incorporate both egocentric and exocentric perspectives.
Majumder et al. \cite{majumder2024viewpoint} utilize language descriptions as weak supervision, as shown in Fig. \ref{fig:joint_sel}. Specifically, the approach generates captions for each view via video captioning models. These captions are scored against ground-truth narration and ranked to produce best-view pseudo-labels, which are utilized to train the view selection model.  
Another work \cite{switch-a-View} proposes a pretext task to detect view switches in instructional videos with varying viewpoints. The model trained for this task is subsequently repurposed to train a view selection model.

\noindent$\bullet~$Discussion: In ego-exo settings, the field-of-view disparity between egocentric (close-up) and exocentric (wide-angle) perspectives poses unique challenges compared to traditional multi-view selection systems. This requires models to determine whether the current task phase demands a focused ``zoomed-in" or a contextual ``far-view" perspective. This challenge is especially important in instructional videos for educational purposes. Future research could integrate user-specific preferences into view selection criteria.

\section{Datasets} \label{section:dataset}
We introduce publicly available datasets offering both egocentric and exocentric perspectives. We categorize these datasets based on domain and describe their intended purposes, views, annotations, and unique features. This overview helps researchers select suitable datasets for their studies.

Table \ref{dataset} provides a summary of the datasets. For datasets that provide synchronized videos, we list the number of first-person and third-person viewpoints. Most datasets cover multiple activity types, while others focus on activities in specialized scenarios. Additionally, datasets \cite{egohumans,jia2020lemma,FT-HID,kong2024wts,Ego2top,Fan_2017_CVPR, Joint_Seg, Complementary_han, bev2, YOWO, hein2023nextgeneration} include multi-agent settings, involving multiple participants in a video. This facilitates the analysis of human interactions and collaboration in complex activities. Furthermore, datasets \cite{huang2024egoexolearn, lbw,kong2024wts,Egoexo4d,ilaslan2023gazevqa, hein2023nextgeneration,Huang_2018predicting} provide egocentric eye gaze information, offering valuable insights into human intention and decision-making process. Below, we provide a detailed description of each dataset.

\textbf{\emph{A. Action Understanding.}}
Most ego-exo action understanding datasets focus on activities in specific scenarios or controlled environments. \textbf{CMU-MMAC} \cite{CMU-MMAC} records videos of individuals cooking recipes in a lab kitchen. \textbf{H2O} \cite{H2O}, \textbf{Assembly101}\cite{sener2022assembly101}, \textbf{ARCTIC}~\cite{arctic} and \textbf{OAKINK2} \cite{Zhan2024OakInk2A} focus on hand-object manipulation on the tabletop.
\textbf{Homage} \cite{homeage} captures daily life activities in two houses. \textbf{LEMMA} \cite{jia2020lemma} features multi-agent goal-directed daily activities in living room and kitchen scenarios. \textbf{FT-HID} \cite{FT-HID} focuses on multi-person interactions and includes 30 human interaction action classes. \textbf{EgoExo-Fitness} \cite{fitness} focuses on full-body action understanding in natural fitness scenarios.
\textbf{Charades-Ego} \cite{actorobserver} leverages scripts from the Charades \cite{Sigurdsson2016HollywoodIH} and self-collected data, recording multi-view videos of participants performing these scripts. \textbf{CORE4D-Real}\cite{Zhang2024CORE4DA4} uniquely captures multi-person and object interactions in household object rearrangement. 

More recent datasets involve diverse activities in multiple environments. \textbf{Ego-Exo4D} \cite{Egoexo4d} is a large-scale multi-view dataset focused on skilled human activities. It offers multimodal annotations, including audio, eye gaze, 3D point clouds, and detailed language descriptions. 
Both \textbf{EgoExoLearn}\cite{huang2024egoexolearn} and \textbf{EgoMe} \cite{EgoMe} include exocentric demonstration videos and corresponding egocentric recordings of individuals performing the tasks based on the demonstrations. These datasets offer valuable resources for studying how humans interpret and adapt actions from an external perspective to their own.

To analyze human motion, \textbf{EgoPW} \cite{egopw}, \textbf{First2Third-Pose} \cite{dhamanaskar2023enhancing} and \textbf{ECHP} \cite{liu2023egofish3d} are designed for egocentric human full-body pose estimation with support from third-person cameras. Specifically, egocentric videos in ECHP~\cite{liu2023egofish3d} are recorded using a head-mounted fisheye camera. \textbf{AssemblyHands} \cite{ohkawa2023assemblyhands} and \textbf{ThermoHands} \cite{Ding2024ThermoHandsAB} focus on hand-object interaction and provide hand pose annotations. \textbf{EgoHumans} \cite{egohumans} features 3D pose estimation and tracking. \textbf{Nymeria} \cite{Nymeria} is a large-scale motion dataset collected in the wild, featuring multimodal egocentric data and a third-person view by an observer. In the surgical domain, \textbf{Hein et al.} \cite{hein2023nextgeneration} propose a multi-view dataset for the pose estimation of surgical instruments. 

\textbf{OVR} \cite{dwibedi2024ovr} is the first multi-view dataset for temporal repetition counting. This task aims to identify repetitive events in a video. Videos in OVR~\cite{dwibedi2024ovr} are sourced from exocentric dataset Kinetics \cite{kinetic} and egocentric dataset Ego4D \cite{ego4d}. Annotations include the start and end times of repetitions, the number of repetitions, and action descriptions. The open-vocabulary semantics of OVR~\cite{dwibedi2024ovr} support text-conditioned repetition counting.

\textbf{\emph{B. Driving.}}
Integrating both in-vehicle and out-vehicle views can provide a comprehensive understanding of the driver’s behavior. 
\textbf{LBW}\cite{lbw} is a multi-view driving dataset for driver's attention estimation. It includes gaze data from eye-tracking glasses and the forward road scene. 
\textbf{AIDE}\cite{yang2023aide} is designed for assistive driving perception, capturing naturalistic driving from four views: three external (front, left, right) and one internal (driver's state). Annotations cover facial expressions, body postures, gestures, and vehicle conditions. 
\textbf{WTS}\cite{kong2024wts} provides not only vehicle and infrastructure perspectives, but also pedestrian perspectives. It can advance fine-grained video event detection.

\textbf{\emph{C. Affordance Grounding.}}
\textbf{AGD20K}\cite{Luo_2022_CVPR} is the earliest image-level multi-view affordance grounding dataset. It classifies the collected data into seen and unseen sets to evaluate the model's generalization ability. It has become a widely used benchmark for numerous methods. 
To advance dexterous manipulation research, \textbf{FAH} \cite{yang2024learning} identifies multi-finger grasping regions through detailed hand movement categorization. 
\textbf{PAD} \cite{luo2021one} provides pixel-level annotations, enabling precise affordance grounding through semantic segmentation models.

\textbf{\emph{D. Generation.}}
\textbf{ThirdtoFirst} \cite{li2021ego} is designed for exocentric to egocentric image synthesis. It consists of 531 temporally aligned video pairs. Video collectors perform various actions in front of the exocentric camera (side or top-view), while a body-worn camera captures their motion from the first-person perspective.

\textbf{\emph{E. Scene Understanding.}}
\textbf{360+x}\cite{chen2024360+} is a multi-view, multi-modal panoptic scene understanding dataset. It includes third-person panoramic and front views, as well as first-person monocular and binocular views. The dataset also offers audio, location data, and textual scene descriptions. Benchmarks include video scene classification, temporal action localization, and cross-modality retrieval.

\textbf{\emph{F. Video Question Answering.}}
\textbf{GazeVQA} \cite{ilaslan2023gazevqa} is designed for task-oriented video question answering. It features collaboration between an instructor and a novice in assembling or disassembling an industrial product. A key feature of GazeVQA \cite{ilaslan2023gazevqa} is the inclusion of egocentric eye gaze information, which aids in understanding human intention.

\textbf{\emph{G. Egocentric Wearer Identification.}}
\textbf{Ego2Top} \cite{Ego2top}, \textbf{IUShareView} \cite{Fan_2017_CVPR} , and \textbf{TF2023} \cite{Joint_Seg2} utilize a fixed exocentric camera and multiple egocentric cameras mounted on different individuals in the environment. In IUShareView \cite{Fan_2017_CVPR} and TF2023 \cite{Joint_Seg2}, each person is annotated with a unique ID. Additionally, TF2023 \cite{Joint_Seg2} provides segmentation masks for individuals in third-person views.

\textbf{\emph{H. Cross-View Human Identification.}}
\textbf{CVMHT} \cite{Complementary_han} comprises over 23K frames of top-view and horizontal-view videos from five different locations. Annotations include bounding boxes and cross-view ID numbers for subjects. \textbf{DMHA} \cite{connect_han} is a synthetic dataset featuring top-view and side-view videos from common outdoor surveillance scenes. Compared to CVMHT\cite{Complementary_han}, it also includes the side-view camera's location and view direction in the top-view.

\begin{table*}[t]
\scriptsize
\caption{Overview of ego-exo datasets: `Data Statistics' shows video/frame/hour stats. `Ego/Exo Views' lists viewpoints for synchronized datasets. `Multi-Activities' indicates varied activities. `Multi-Agents' denotes interactions among multiple people.}
\label{dataset}
\centering
\renewcommand{\arraystretch}{1.1}
\begin{tabular}{l|c|c|c|c|c|c|c|c}
\hline
\rowcolor{gray!15}
\textbf{Dataset}                & \textbf{Year} & \textbf{Domain}                  & \textbf{Data Statistics}      & \textbf{Exo Views} & \textbf{Ego Views} & \textbf{Multi-Activities}   & \textbf{Multi-Agents}      & \textbf{Gaze}      \\ \hline
CMU-MMAC\cite{CMU-MMAC}         & 2008 & Action Understanding             & 1050 videos       & 3        & 2        & \ding{55} & \ding{55}    & \ding{55} \\ 
Charades-Ego\cite{actorobserver}& 2018 & Action Understanding             & 7.4M frames            & 1        & 1        & \ding{51} & \ding{55}    & \ding{55} \\
LEMMA\cite{jia2020lemma}        & 2020 & Action Understanding             & 4.1M frames         & 2        & 1        & \ding{51} & \ding{51}    & \ding{55} \\
H2O\cite{H2O}                   & 2021 & Action Understanding             & 571K frames            & 4        & 1        & \ding{55} & \ding{55}    & \ding{55} \\ 
HOMAGE\cite{homeage}            & 2021 & Action Understanding             & 25.5 hours         & 1-4      & 1        & \ding{51} & \ding{55}    & \ding{55} \\
Assembly101\cite{sener2022assembly101}    & 2022 & Action Understanding   & 110M frames          & 8        & 4        & \ding{55} & \ding{55}    & \ding{55} \\
EgoPW\cite{egopw}               & 2022 & Action Understanding             & 318K frames   & 1         & 1         & \ding{51} & \ding{55}     & \ding{55}     \\ 
ARCTIC\cite{arctic}             & 2023 & Action Understanding             & 2.1M frames       & 8        & 1        & \ding{55} & \ding{55}    & \ding{55} \\ 
FT-HID\cite{FT-HID}             & 2023 & Action Understanding             & 6.4M frames      & 3        & 2        & \ding{51} & \ding{51}    & \ding{55} \\ 
EgoHumans\cite{egohumans}      & 2023  & Action Understanding             & 571K frames    & 8-15      & 1         & \ding{51} & \ding{51}     & \ding{55}      \\ 
AssemblyHands\cite{ohkawa2023assemblyhands}     & 2023 & Action Understanding  & 3.03M frames    & 8         & 4         & \ding{55} & \ding{55}     & \ding{55}    \\
First2Third-Pose\cite{dhamanaskar2023enhancing} & 2023 & Action Understanding  & 190K frames      & 2-3       & 1         & \ding{51} & \ding{55}     & \ding{55}     \\ 
ECHP\cite{liu2023egofish3d}     & 2023 & Action Understanding             & 75K frames     & 2         & 1         & \ding{51} & \ding{55}     & \ding{55}     \\ 
Hein et al. \cite{hein2023nextgeneration}  & 2023 & Action Understanding  & 1.7M frames    & 5         & 2         & \ding{55} & \ding{51}     & \ding{51}     \\
OAKINK2\cite{Zhan2024OakInk2A}  & 2024 & Action Understanding             & 4.01M frames       & 3        & 1        & \ding{51} & \ding{55}    & \ding{55} \\
EgoExo-Fitness\cite{fitness}    & 2024 & Action Understanding             & 1276 videos           & 3        & 3        & \ding{51} & \ding{55}    & \ding{55} \\ 
CORE4D-Real\cite{Zhang2024CORE4DA4}  & 2024 & Action Understanding        & 1K videos         & 4        & 1        & \ding{55} & \ding{51}    & \ding{55} \\
Ego-Exo4D\cite{Egoexo4d}        & 2024 & Action Understanding             & 1286 hours         & 4        & 1        & \ding{51} & \ding{55}    & \ding{51} \\ 
EgoExoLearn\cite{huang2024egoexolearn}    & 2024 & Action Understanding   & 120 hours        & -        & -        & \ding{51} & \ding{55}    & \ding{51} \\ 
ThermoHands\cite{Ding2024ThermoHandsAB}     & 2024 & Action Understanding  & 96K frames    & 1         & 1         & \ding{51} & \ding{55}     & \ding{55}          \\ 
Nymeria\cite{Nymeria}           & 2024 & Action Understanding     &  201M frames       & 1        & 1         & \ding{51} & \ding{55}     & \ding{51} \\
OVR\cite{dwibedi2024ovr}        & 2024 & Action Understanding     & 72552 videos     & -         & -         & \ding{51} & \ding{55}     & \ding{55} \\
EgoMe\cite{EgoMe}               & 2025 & Action Understanding     & 15804 videos      & 1        & 1         & \ding{51} & \ding{55}     & \ding{51} \\ \hline

LBW\cite{lbw}                   & 2022 & Driving                          &   123K frames      & 1         & 2         & \ding{55} & \ding{55}     & \ding{51} \\ 
AIDE\cite{yang2023aide}         & 2023 & Driving                          &   521.6K frames  & 1         & 3         & \ding{55} & \ding{55}     & \ding{55}  \\ 
WTS\cite{kong2024wts}           & 2024 & Driving                          &   52.8K frames   & 18        & 2         & \ding{55} & \ding{51}     & \ding{51} \\ \hline

PAD\cite{luo2021one}            & 2021 & Affordance Grounding             &  4K frames     & 1        & 1          & -         & -             & - \\ 
AGD20K\cite{Luo_2022_CVPR}      & 2022 & Affordance Grounding             &  20K frames     & 1        & 1          & -         & -             & - \\
FAH\cite{yang2024learning}      & 2024 & Affordance Grounding             &  6K frames     & 1        & 1          & -         & -             & - \\ \hline

Thirdtofirst\cite{li2021ego}    & 2021 & Generation                       &  334.6K frames & 1        & 1          & \ding{51} & \ding{55}     & \ding{55} \\ \hline

360+x\cite{chen2024360+}        & 2024 & Scene Understanding              &  8.5M frames     & 2         & 2         & \ding{51} & \ding{55}     & \ding{55} \\ \hline

GazeVQA\cite{ilaslan2023gazevqa} & 2023 & Video Question Answering         & 125 hours   & 2        & 1          & \ding{55} & \ding{55}     & \ding{51} \\ \hline

Ego2Top\cite{Ego2top}               & 2016 & Egocentric Wearer Identification          & 225K frames       & 1      & 1-6     & \ding{51} & \ding{51} & \ding{55} \\
IUShareView\cite{Fan_2017_CVPR}     & 2017 & Egocentric Wearer Identification          & 11.2K frames    & 1      & 2       & \ding{51} & \ding{51} & \ding{55}    \\  
TF2023\cite{Joint_Seg2}             & 2024 & Egocentric Wearer Identification          & 49.8K frames    & 1      & 2       & \ding{51} & \ding{51} & \ding{55}   \\ \hline  

CVMHT\cite{Complementary_han}       & 2020 & Cross-View Human Identification &  23K frames        & 1      & 2-3     & \ding{51} & \ding{51} & \ding{55} \\  
DMHA \cite{connect_han}             & 2022 & Cross-View Human Identification &  84.8K frames     & 1      & 1       & \ding{55} & \ding{51} & \ding{55} \\ \hline

CSRD-II\cite{bev2}              & 2022 & Camera Registration &  2K frames   & 1     & 2 & \ding{51} & \ding{51} & \ding{55}  \\   
CSRD-V\cite{bev2}               & 2022 & Camera Registration &  5K frames  & 1     & 5 & \ding{51} & \ding{51} & \ding{55}  \\  
YOWO\cite{YOWO}                 & 2024 & Camera Registration &  -    & 5-17  & 1 & -         & -         & -          \\ \hline

\end{tabular}
\vspace{-5mm}
\end{table*}

\textbf{\emph{I. Camera Localization.}}
\textbf{CSRD-II} \cite{bev2} and \textbf{CSRD-V} \cite{bev2} are synthetic datasets for egocentric camera localization. Annotations include subject positions and camera poses in the bird's-eye view. \textbf{YOWO} \cite{YOWO} is a synthetic dataset for exocentric camera localization. An agent with an egocentric camera traverses the scene, collaborating with ceiling-mounted cameras for scene reconstruction and camera localization.

\section{Discussion} \label{section:outlook}
This section discusses the limitations of current research and offers insights into future directions from the perspectives of data, model, and application. 

\noindent\textbf{Insights from Data.}
Most existing datasets focus on daily life activities, resulting in a scarcity of data tailored to specific scenarios such as public service, healthcare, and education. This limitation hinders the development of approaches for specialized applications. Additionally, most datasets use sophisticated multi-camera setups to record synchronized egocentric and exocentric videos. This significantly increases costs and limits the scalability of data collection. Future research could investigate transforming existing unpaired egocentric~\cite{ego4d,epic} and exocentric~\cite{kinetic,ava,Soomro2012UCF101AD} datasets to enable collaboration between these perspectives. Furthermore, integrating video data with other modalities, such as audio \cite{jia2024audio} and IMU sensors \cite{zhang2025masked}, could enrich the captured information, providing a more comprehensive understanding of complex scenarios. 

\noindent\textbf{Insights from Model.}
Most existing models are designed for specific tasks and lack generalizability. In contrast, recent advancements in vision-language models (VLMs) \cite{llava,internvl,li2025eagle2,chen2023videollm} highlight their effectiveness to handle diverse tasks. Future research could explore equipping VLMs with the capability to integrate egocentric and exocentric perspectives, facilitating unified cross-view tasks in a single framework. Moreover, current methods often rely on synchronized egocentric and exocentric data. However, the limited scale of such paired datasets hinders the effective training of large models. To overcome this limitation, promising directions include leveraging alignment strategies or retrieval-augmented methods \cite{videorag} to better utilize unpaired data.

\textbf{Insights from Application.}
Current research are primarily centered on daily life contexts, with limited attention to specialized application domains. For instance, while affordance grounding has been well-studied for everyday objects~\cite{Luo_2022_CVPR, Li_2023_CVPR,xu2024weakly,zhang2024self,Rai_2024_CVPR}, predicting affordance regions for surgical tools or industrial components receives less attention. Extending egocentric and exocentric collaboration techniques to domains such as medicine and industry could unlock new opportunities in these fields.

\section{Conclusion}
This survey presents a comprehensive review of cross-view collaboration with egocentric and exocentric vision. We begin by discussing the practical value of egocentric and exocentric collaboration across various applications. We then link these applications to key research tasks required to realize them. Current research advancements are categorized into three directions: egocentric for exocentric, exocentric for egocentric, and joint learning, with a detailed overview of progress in each area. In addition, we review relevant datasets that support both perspectives. Finally, we provide a discussion on data, models, and applications, and outline future research directions. We hope this review inspires deeper exploration into egocentric-exocentric collaboration, paving the way for artificial intelligence to perceive the world with human-like vision.

{\small
\bibliographystyle{IEEEtran}
\bibliography{IEEEabrv,extend_IEEEabrv, mylib}
}

\end{document}